%% file: main.tex
\documentclass[multiauthors,onecolumn,cfinal]{LaTeXclass/cohere}
\usepackage{ragged2e}
\author{name={My Chiffon Nguyen\fa},affiliation={1,2}}
\author{name={Tianyi Hu\fa},affiliation={3}}
\author{name={Vinod Anbalagan\fa},affiliation={4,1}}
\author{name={Henry Gagnier\fa},affiliation={5,1}}
\author{name={Pouya Sadeghi\fa},affiliation={6,1}}
\author{name={Farah Abdou},affiliation={7}}
\author{name={Aston C Garratt},affiliation={8}}
\author{name={Diego MMS Oliveira},affiliation={8}}
\author{name={Tidor-Vlad Pricope},affiliation={8}}
\author{name={Isaac Saxton-Knight},affiliation={8}}
% ...
\affiliations{
    \item[1] Cohere Labs Community
    \item[2] Lida Safety
    \item[3] Aarhus University
    \item[4] University of Toronto
    \item[5] University of Rochester
    \item[6] University of Waterloo
    \item[7] Cambridge Centre for Alternative Finance
    \item[8] Poindexter Labs
    \item[9] Cohere Labs
    \item[10] Obafemi Awolowo University, Ile-Ife
    \item[11] German Research Center for Artificial Intelligence (DFKI)
    \item[12] Saarland University
    \item[13] Andhra University
    \item[14] École Polytechnique
    \item[15] University of Copenhagen
    \item[16] University of British Columbia
    \item[17] Nile University
    \item[18] Adobe Systems
    \item[19] Crane AI Labs
    \item[20] RV College of Engineering
    \item[21] FPT University
    \item[22] Lexsi Labs
    \item[23] University of St. Gallen
    \item[24] AI Singapore
    \item[25] Indian Institute of Technology, Kharagpur
    \item[26] University of Hull
    \item[27] 42 Abu Dhabi
    \item[28] Queen Mary University of London
    \item[29] Tribhuvan University
    \item[30] Universitas Muslim Indonesia
    \item[31] Persistent Systems
    \item[32] Johns Hopkins University
    \item[33] University of Michigan-Dearborn
    \item[34] Wichita State University
    \item[35] Baylor University
    \item[36] ETH Zurich
    \item[37] The Hong Kong University of Science and Technology (Guangzhou)
    \item[38] Singapore University of Technology and Design
    \item[39] Agency for Science, Technology and Research (A*STAR)
    \item[40] University of Ibadan
    \item[41] University of Vienna
    \item[42] Oracle
    \item[43] Fondazione Bruno Kessler
    \item[44] Imperial College London
    \item[45] Masakhane Community
    \item[46] Georgia Institute of Technology
    \item[47] Independent
    \item[48] University of Belgrade
    \item[49] Universidade Federal Fluminense
    \item[50] Universidad Nacional de Tres de Febrero
    \item[51] New York University
    \item[52] Mila - Quebec Artificial Intelligence Institute
    \item[53] Addis Ababa University
    \item[54] Carnegie Mellon University
    \item[55] Universitat Politècnica de Catalunya
    \item[56] Barcelona Supercomputing Center
    \item[57] Indian Institute of Science
    \item[58] Universidade de Sorocaba
    \item[59] Platned
    \item[60] University of Michigan - Ann Arbor
    \item[61] Vivoka
    \item[62] Conservatoire National des Arts et Métiers (Cnam)
    \item[63] University of Amsterdam
    \item[64] University of Cincinnati
    \item[65] University of Helsinki
    \item[66] University of Udine
    \item[67] Universidade de São Paulo
    \item[68] Stanford University
    \item[69] Harvard University
    \item[70] Massachusetts Institute of Technology
    \item[71] Minia University
    \item[72] Egypt-Japan University of Science and Technology
    \item[73] University of Minnesota - Twin Cities
    \item[74] Pune Vidyarthi Griha's College of Engineering, Technology and Management
    \item[75] University of Zurich
    \item[76] NFQ Advisory Services
    \item[77] University of Maryland, College Park
    \item[78] Stevens Institute of Technology
    \item[79] Maastricht University
    \item[80] Sabrish Institute of French, Bangalore
    \item[81] University of California, Los Angeles
    \item[82] University of Cambridge
    \item[83] George Mason University
    \item[84] Luxembourg Institute of Science and Technology
    \item[85] Universidad del Pacífico
    \item[86] Heriot-Watt University
    \item[87] Rochester Institute of Technology
    \item[88] Fundação Getúlio Vargas
    \item[89] Jahangirnagar University
    \item[90] Center for Low-Resource Languages and Cultures
    \item[91] Emory University
    \item[92] Ukrainian Catholic University
    \item[93] University of Luxembourg
    \item[94] St. Xavier's College, Jawalakhel
    \item[95] Universidad Andina del Cusco
    \item[96] Université Pierre et Marie Curie
    \item[97] Sorbonne Université
    \item[98] Universitas Indonesia
    \item[99] University of Maryland, Baltimore County
    \item[100] University College London
    \item[101] Telefónica Research
    \item[102] Oregon State University
    \item[103] Indian Institute of Technology, Hyderabad
    \item[104] Indian Institute of Technology, Roorkee
    \item[105] Boston University
    \item[106] École de technologie supérieure
    \item[107] Fufong Labs
    \item[108] University of Texas at Dallas
    \item[109] Clemson University
}

\corresponding[*]{
\url{hi@mychiffonn.com},
\url{tenney.hu@cs.au.dk},
\url{vinodanbalagan@gmail.com},
\url{henrygagnier9@gmail.com},
\url{pouya.sadeghi@uwaterloo.ca},
\url{akhil.arora@cs.au.dk},
\url{juliakreutzer@cohere.com}}

\usepackage{url}
\usepackage{booktabs}
\usepackage{graphicx}
\usepackage{capt-of}
\usepackage{multirow}
\usepackage{array,booktabs,tabularx,threeparttable,pifont}
\usepackage[nopatch]{microtype}
\usepackage{inconsolata}
\usepackage{graphicx}
\usepackage[table]{xcolor}
\usepackage[most]{tcolorbox}
\tcbuselibrary{skins,breakable}

\definecolor{tzBlueHeader2}{RGB}{105,185,225}
\definecolor{tzBlueBorder}{RGB}{115,190,225}
\definecolor{tzBlueFill}{RGB}{232,246,252}

\usepackage{titlesec}
\titleformat{\paragraph}[runin]{\normalfont\bfseries}{}{0pt}{}

\titlespacing*{\paragraph}{0pt}{3.25ex plus 1ex minus .2ex}{1em}

\usepackage{enumitem} 

\usepackage{wrapfig} 

\newtcolorbox{jsonbox}{enhanced, colback=gray!6!white, colframe=gray!55!black,
  boxrule=0.5pt, arc=2pt, left=6pt, right=6pt, top=4pt, bottom=4pt}

\usepackage{xspace}
\tcbuselibrary{skins,listings,breakable}

\newtcblisting{promptbox}[1][]{
  enhanced,
  breakable,
  width=\linewidth,
  colback=tzBlueFill,
  colframe=tzBlueBorder,
  boxrule=1.2pt,
  arc=6pt,
  left=5pt, right=5pt, top=4pt, bottom=2pt,
  fonttitle=\bfseries\small,
  coltitle=white,
  colbacktitle=tzBlueHeader2,
  title={#1},
  listing only, unbreakable,
  listing options={
    basicstyle=\ttfamily\scriptsize,
    breaklines=true,
    breakatwhitespace=true,
    breakindent=1.5em,
    columns=fullflexible,
    keepspaces=true,
    showstringspaces=false
  }
}

\newtcolorbox{riddleexample}[1][]{
  enhanced,
  colback=orange!4!white,
  colframe=orange!60!black!50,
  boxrule=0.6pt,
  arc=3pt,
  left=8pt, right=8pt, top=6pt, bottom=6pt,
  fonttitle=\bfseries\small\sffamily,
  coltitle=white,
  colbacktitle=orange!60!black!70,
  title={#1},
  attach boxed title to top left={yshift=-2mm, xshift=4mm},
  boxed title style={arc=2pt, boxrule=0pt, left=6pt, right=6pt},
}

\usepackage{cleveref}
\crefname{appendix}{App.}{Apps.}
\Crefname{appendix}{App.}{Apps.}
\usepackage{placeins}

\definecolor{dominantrow}{HTML}{87BFE0} 
\newcommand{\dominantrow}{\rowcolor{dominantrow}}
\newcommand{\dataset}{\textsc{CulturalRiddles}\xspace}

\usepackage{fontspec}
\usepackage{polyglossia}
\setdefaultlanguage{english}
\setotherlanguage{arabic}
\AtBeginDocument{}

\newfontfamily\arabicfont{Amiri}[
  Path=fonts/, Script=Arabic, Scale=1.1, Extension=.ttf,
  UprightFont=*-Regular, BoldFont=*-Bold,
  ItalicFont=*-Italic, BoldItalicFont=*-BoldItalic
]

\newcommand{\arabtext}[1]{\textarabic{#1}}
\newcommand{\arabword}[1]{{\arabicfont\textarabic{#1}}}

\newcommand{\cmark}{\textcolor{green!50!black}{\ding{51}}}
\newcommand{\xmark}{\ding{55}}

\newcommand{\numcommunities}{62\xspace}
\newcommand{\numlang}{51\xspace}
\newcommand{\numcountries}{45\xspace}
\newcommand{\numriddles}{6,191\xspace}
\newcommand{\numbenchmarkedriddles}{6,090\xspace}
\newcommand{\numbenchmarkedcommunities}{61\xspace}
\newcommand{\nummodels}{28\xspace}
\newcommand{\numresponses}{170,520\xspace}
\newcommand{\numhumanresponses}{100,769\xspace}
\newcommand{\numuniquehumanresponses}{68,067\xspace}

\newcommand{\invalidhumanpercent}{2.38\%\xspace}
\newcommand{\numhumanscoredcommunities}{57\xspace}
\newcommand{\numhumanevaluatedlanguages}{48\xspace}
\newcommand{\numiaacommunities}{19\xspace}

\newcommand{\humanevalpercent}{59.10\%\xspace}

\newcommand{\autoresolvedpercent}{18.87\%\xspace}
\newcommand{\exactmatchpercent}{16.45\%\xspace}

\newcommand{\judgedpercent}{91.93\%\xspace}

\newcommand{\unscoredpercent}{1.78\%\xspace}

\newcommand{\numerrortyperesponses}{132,940\xspace}
\newcommand{\declinedpercent}{0.80\%\xspace}
\newcommand{\declinedmaxpercent}{3.81\%\xspace}

\newcommand{\hallucinationpercent}{24.19\%\xspace}

\newcommand{\correctvariantpercent}{9.51\%\xspace}
\newcommand{\differentvalidpercent}{6.16\%\xspace}

\newcommand{\jurykappahigh}{0.92\xspace}
\newcommand{\jurykappaextremelylow}{0.88\xspace}
\newcommand{\jurykappamin}{0.65\xspace}
\newcommand{\juryminlanguage}{Swahili\xspace}
\newcommand{\jurykappamax}{0.98\xspace}
\newcommand{\jurymaxlanguage}{Belarusian and Catalan\xspace}
\newcommand{\juryuncertaincommunities}{Swahili (Kenya)\xspace}

\newcommand{\numcleanediaapairs}{4,246\xspace}
\newcommand{\numcleanediaariddles}{1,074\xspace}
\newcommand{\numsettingapairs}{4,018\xspace}
\newcommand{\numsettingariddles}{1,028\xspace}
\newcommand{\numsettingbpairs}{25,370\xspace}
\newcommand{\numsettingcpairs}{33,880\xspace}
\newcommand{\poolaccuracyhigh}{37.0\%\xspace}
\newcommand{\poolaccuracymedium}{30.8\%\xspace}
\newcommand{\poolaccuracylow}{18.0\%\xspace}
\newcommand{\poolaccuracyextremelylow}{12.7\%\xspace}
\newcommand{\resourcesdmin}{8\xspace}
\newcommand{\resourcesdmax}{13\xspace}

\newcommand{\formlugandawhoamipercent}{98\%\xspace}
\newcommand{\formjapanesewhatisitpercent}{98\%\xspace}
\newcommand{\numenglishriddles}{499\xspace}
\newcommand{\numenglishrevealing}{41\xspace}
\newcommand{\numenglishrevealinglumasaba}{38\xspace}
\newcommand{\numenglishrevealingsudanese}{3\xspace}
\newcommand{\numenglishkept}{458\xspace}
\newcommand{\englishoriginalaccuracy}{22.1\%\xspace}
\newcommand{\englishaccuracy}{32.4\%\xspace}
\newcommand{\englishgain}{10.3\xspace}
\newcommand{\englishgainlow}{7.6\xspace}
\newcommand{\englishgainhigh}{13.0\xspace}
\newcommand{\englishhighresourceaccuracy}{69\%\xspace}
\newcommand{\numsettingbsampled}{25,372\xspace}
\newcommand{\settingbeffective}{17,032\xspace}
\newcommand{\numsettingbraisedcommunities}{33\xspace}
\newcommand{\settingbtokensgemini}{60.8M/10.5M\xspace}
\newcommand{\settingbtokensdeepseek}{71.6M/5.1M\xspace}
\newcommand{\bestmodelaccuracy}{79.4\%\xspace}
\newcommand{\numwriters}{167\xspace}
\newcommand{\numreviewers}{109\xspace}

\newcommand{\numsurveyrespondents}{76\xspace}
\newcommand{\numsurveycommunities}{43\xspace}

\title{\dataset}
\subtitle{A Multicultural Benchmark of Riddles}

\abstract{
As LLMs expand their language coverage, evaluating how well they understand and represent diverse cultures becomes increasingly important. However, many benchmarks remain English-centric, with limited cultural coverage and a focus on factual knowledge over its application in real-world cultural contexts. We propose \dataset\footnote{\url{https://huggingface.co/datasets/CohereLabsCommunity/CulturalRiddles}}, a benchmark of original, culturally grounded riddles created entirely by humans through a participatory community effort, spanning \numcommunities cultures (\numcountries countries, \numlang languages). These riddles are deeply rooted in their respective cultures and require abductive and compositional reasoning, alongside detailed factual and social knowledge, language skills, and an understanding of culturally specific associations. We benchmark \nummodels LLM configurations using both automatic and large-scale human evaluation. Even frontier models achieve at most 22\% accuracy in some communities. We publicly release all artifacts alongside an open contribution process to expand the benchmark's linguistic and cultural coverage\footnote{\url{https://github.com/Cohere-Labs-Community/CulturalRiddles}}.
}

\begin{document}

\section{Introduction}

As Large Language Models (LLMs) are deployed worldwide, they must operate not only across languages but also across cultural contexts~\citep{hershcovich2022challenges, singh2025global, ustun2024aya}.
Recent LLMs increasingly overcome language barriers~\citep{ghosh2025survey, salamanca2026tiny}, but whether and when they adequately model cultures around the globe remains an open question~\citep{myung2024blend, naous2024having, chiu2025culturalbench, rao2025normad, chang2025global, khan-etal-2026-lie}.

\begin{figure}
    \centering
    \includegraphics[width=\linewidth]{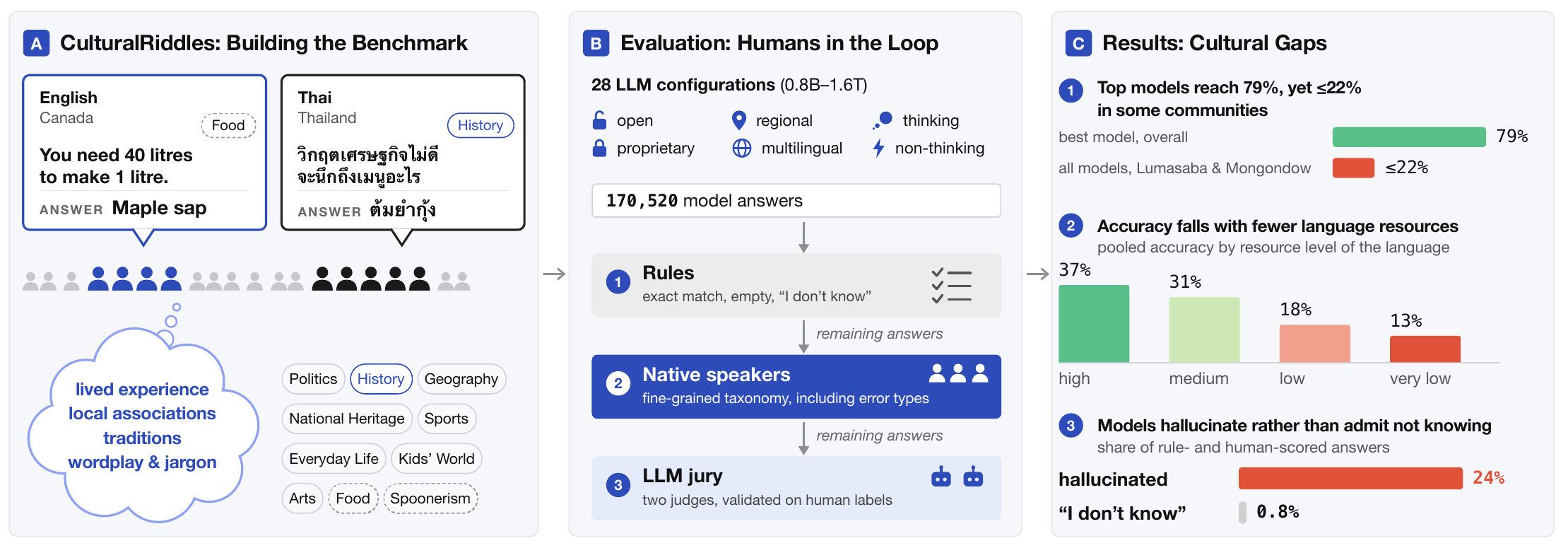}
    \caption{Overview of the \dataset benchmark, the LLM evaluation stage and the core results.}
    \label{fig:overview}
\end{figure}

Most cultural benchmarks assess whether models \emph{possess} cultural knowledge~\citep{hu2024bridging, liu-etal-2025-culturally, alkhamissi2026hire, zhou2025culture, chowdhury2025facts}. However, cultural understanding also requires \emph{applying} it to infer implicit meanings from contextual cues, and as LLMs gain greater knowledge and stronger reasoning, their failures increasingly come from choosing the wrong interpretation rather than lacking the fact~\citep{gekhman2026thinking}.
Riddles could test this ability, particularly \textit{compositional and abductive reasoning}~\citep{halder2026riddlebench}. They have long served to test cultural understanding in traditions from ancient Greece, Rome, and India to China~\citep{szalc2011alexander,tan2016solving,monda2016beyond}, and they span topics from history and politics to sports and literature~\citep{kaivola2018riddle}.
Solving a \textit{cultural riddle} requires weighing alternative cultural associations for each clue, combining them into a coherent whole (\textit{compositional reasoning}), and inferring the answer that best explains all clues (\textit{abductive reasoning})~\citep{halder2026riddlebench}.
E.g., as shown in \Cref{tab:examples}, the German riddle requires interpreting ``\emph{the German 9/11}'' in German day--month order, then recalling the historical events and name of 9 November. By contrast, simply asking ``\emph{What is 9 November called in Germany?}'' would require only factual recall. Cultural riddles thus test whether LLMs can interpret cultural clues and combine knowledge with reasoning.

\begin{table*}[t]
    \centering
    \small
    \begin{tabularx}{\linewidth}{@{}l l >{\raggedright\arraybackslash}X >{\raggedright\arraybackslash}X >{\raggedright\arraybackslash}p{3.1cm}@{}}
    \toprule
      \textbf{Community} & \textbf{Topic }& \textbf{Riddle} & \textbf{Riddle (en)} & \textbf{Answer (+en)}\\
       \midrule
        deu (DEU) & History & Name des deutschen 9/11. & Name of the German 9/11. & Schicksalstag (``day of fate'') \\
        \midrule
        ltz (LUX) & Politics & Déi schwaarz Partei. & The black party. & CSV (Christian Social People's Party) \\
        \midrule
        apd (SDN) & Arts & \arabtext{انا السودانية التي لقبت بظبية المسالمة من اكون؟} & I am the Sudanese woman who was nicknamed `The Gazelle of Al-Masalma'. Who am I? & \arabword{عزيزة آدم منديل}\newline (Aziza Adam Mandil) \\
    \bottomrule
    \end{tabularx}
    
    \caption{\textbf{Example riddles from three communities.} \emph{Community} is given by the language code and, in parentheses, the country code. The Sudanese Arabic riddle (third) comes from one of the hardest communities (\Cref{sec:results-cultures}).}

    \label{tab:examples}
\end{table*}



Cultural riddles have been benchmarked for a few languages~\citep{fabbri2025multinrc, boczka2026hungarianriddles, sayeedi2026can, minggad-etal-2026-monculture}, where they challenge even frontier LLMs. However, existing cultural benchmarks
have three key limitations: \textbf{(L1)} narrow linguistic and geographic coverage, \textbf{(L2)} insufficient validation of data quality and difficulty, and \textbf{(L3)} the lack of a robust, scalable, human-informed evaluation protocol. Prior cultural QA work either human-scores a small subset of answers or uses automatic judges~\citep{arora2025calmqa,hasan2025nativqa,libovicky2026cusqa}.

We address these limitations with \textbf{\dataset}, a benchmark of \numriddles original riddles written by community members in their own languages. \dataset spans \numcommunities communities in \numlang languages and \numcountries countries. It includes languages absent from prior cultural benchmarks, and many-to-many links between languages and countries, such as seven languages in multiple countries and seven countries with multiple languages. (\textbf{L1}, \Cref{fig:map}). Native speakers other than the author verify every riddle (\textbf{L2}). We benchmark \nummodels proprietary and open model configurations and, to our knowledge, conduct the largest and most multilingual human evaluation of model answers to open-ended cultural knowledge tasks: native speakers labeled \numuniquehumanresponses unique answers across \numhumanscoredcommunities communities in \numhumanevaluatedlanguages languages, covering \humanevalpercent of \numresponses model responses; rule-based scoring and two human-validated LLM judges score most remaining responses, with unresolved cases excluded from accuracy (\textbf{L3}).


Our results reveal substantial cultural gaps: the best model achieves \bestmodelaccuracy overall accuracy, yet no model exceeds 22\% in the \textit{Lumasaba} and \textit{Mongondow} communities. Translating riddles from five of the hardest communities into English closes only about one-fifth of the accuracy gap to high-resource communities, suggesting that language barriers alone do not explain these differences. We also find that exact-match scoring misses many valid answers and alters model rankings, highlighting the importance of human-informed evaluation.

Through riddles written and reviewed by community members, \dataset offers a nuanced assessment of \textit{how well current models can solve original, culturally grounded riddles in their respective languages}, probing lived experience and locally specific knowledge beyond well-documented facts. The benchmark can also grow to include new communities and track model progress over time.






\section{Related Work}\label{sec:related}

\begin{table*}[t]
\centering
\footnotesize
\setlength{\tabcolsep}{2.5pt}
\renewcommand{\arraystretch}{1.0}

\begin{tabularx}{\linewidth}{
    @{}l
    l
    c
    c
    c
    c
    r
    r
    >{\raggedright\arraybackslash}X@{}
}
\toprule
\textbf{Benchmark}
& \textbf{Tasks}
& \shortstack{\textbf{Open-}\\\textbf{ended}}
& \shortstack{\textbf{Native-}\\\textbf{authored}}
& \shortstack{\textbf{Human-}\\\textbf{only}}
& \shortstack{\textbf{Human}\\\textbf{eval.}}
& \textbf{Lang.}
& \textbf{Locale}
& \textbf{Map} \\
\midrule

\textbf{BLEnD}~\citep{myung2024blend}
& Knowledge
& $\triangle$
& $\triangle$
& \cmark
& \cmark$^\ddagger$
& 13 & 16
& $\sim 1{:}1$ \\

\textbf{CulturalBench}~\citep{chiu2025culturalbench}
& Knowledge
& \xmark
& \cmark
& \xmark
& \xmark
& 1 & 45
& Locale \\

\textbf{INCLUDE}~\citep{romanou2025include}
& Knowledge
& \xmark
& \cmark
& \cmark
& \xmark
& 44 & 52
& $\sim 1{:}1$ \\

\textbf{Global-MMLU}~\citep{singh2025global}
& Knowledge
& \xmark
& \xmark
& \xmark
& \xmark
& 42 & --
& Language \\

\textbf{Global PIQA}~\citep{chang2025global}
& Commonsense
& \xmark
& $\triangle$
& \xmark
& \xmark
& 141$^\dagger$ & --
& -- \\

\textbf{CulturALL}~\citep{lin2026culturall}
& Grounded QA
& $\triangle$
& $\triangle$
& \xmark
& \xmark
& 14 & 51
& Sparse \\

\textbf{Macaron}~\citep{elsetohy2026macaron}
& Reasoning
& \xmark
& \cmark
& \cmark
& \xmark
& 17
& 20
& $\sim 1{:}1$\\

\textbf{CultureForest}~\citep{ye2026cultureforest}
& Norm Reasoning
& $\triangle$
& \xmark
& \xmark
& \xmark
& 1
& 53
& Locale\\

\textbf{CaLMQA}~\citep{arora2025calmqa}
& Long-form QA
& \cmark & $\triangle$ & \xmark & \cmark$^\ddagger$
& 23 & -- & Language \\

\textbf{NativQA}~\citep{hasan2025nativqa}
& Knowledge
& \cmark & \cmark & \cmark & \cmark$^\ddagger$
& 7 & 9 & Sparse \\

\textbf{CUS-QA}~\citep{libovicky2026cusqa}
& Knowledge
& \cmark & $\triangle$ & \xmark & \cmark$^\ddagger$
& 3 & 3 & $\sim 1{:}1$ \\

\midrule
\textbf{MultiNRC}~\citep{fabbri2025multinrc}
& Reasoning
& \cmark & \cmark & \cmark & \xmark
& 3 & -- & Language \\

\textbf{BanglaRiddleEval}~\citep{sayeedi2026can}
& Riddles
& $\triangle$ & \cmark & \xmark & \xmark
& 1 & -- & Language \\

\textbf{MonCulture-eval}~\citep{minggad-etal-2026-monculture}
& Knowledge
& $\triangle$ & \cmark & \cmark & \cmark$^\ddagger$
& 1 & 1 & $\sim 1{:}1$ \\

\textbf{HungarianRiddles}~\citep{boczka2026hungarianriddles} & Riddles & \cmark & \cmark & \cmark & \cmark & 1 & 1 & -- \\

\midrule
\shortstack[l]{\textbf{\dataset} \textbf{(Ours)}}
& Riddles
& \cmark
& \cmark
& $\triangle$
& \cmark
& \numlang & \numcommunities
& \textbf{M:M} \\

\bottomrule
\end{tabularx}

\caption{\textbf{Comparison of cultural and riddle benchmarks.}
\cmark = yes; \xmark = no; $\triangle$ = mixed / partial; ``--'' = not reported. \textbf{Native-authored}: items originated in the target language rather than being translated; \textbf{Human-only}: no LLM-generated benchmark item content; \textbf{Human eval.}: humans score model answers. Locales count distinct (language, country) pairs; Map gives how languages relate to locales, where M:M denotes multiple languages per locale and multiple locales per language. $^\dagger$Language varieties. $^\ddagger$Small-scale human evaluation (\S~\ref{sec:related}). \textbf{\dataset} is the \textbf{only} benchmark that offers large-scale open-ended, community-authored native content and human evaluation with multiple countries per language and multiple languages per country.}
\label{tab:benchmark_comparison}

\end{table*}

\paragraph{Cultural Benchmarks}
Cultural benchmarks for LLMs mostly operationalize culture as knowledge, norms and values, social dynamics, or bias~\citep{adilazuarda-etal-2024-towards,alkhamissi2026hire}. For example, BLEnD~\citep{myung2024blend}, CulturalBench~\citep{chiu2025culturalbench}, and CVQA~\citep{romero2024cvqa} test everyday cultural knowledge, while NormAd~\citep{rao2025normad} and WorldValuesBench~\citep{zhao2024worldvaluesbench} probe social norms and values. These benchmarks are mostly closed-form and test recognition or recall, which says little about whether models apply cultural knowledge when reasoning~\citep{zhou2025culture}. Global PIQA~\citep{chang2025global}, CulturALL~\citep{lin2026culturall}, and Macaron~\citep{elsetohy2026macaron} combine cultural knowledge with multi-step reasoning, and CultureForest~\citep{ye2026cultureforest} requires models to apply culturally grounded norms. While some cultural QA benchmarks include human evaluation, they do so at small scale (120--700 answers evaluated across $<$10 languages)
~\citep{myung2024blend,arora2025calmqa,hasan2025nativqa,libovicky2026cusqa}. Overall, existing benchmarks vary widely in coverage and in how much content is written and evaluated by humans (\Cref{tab:benchmark_comparison}).

\paragraph{Riddle Benchmarks}
General riddle benchmarks test commonsense and lateral thinking predominantly in English: RiddleSense~\citep{lin-etal-2021-riddlesense}, BrainTeaser~\citep{jiang-etal-2023-brainteaser}, and BiRdQA~\citep{zhang2022birdqa} use multiple-choice riddles and puzzles, and RiddleBench~\citep{halder2026riddlebench} collects logic puzzles from exam papers; Chinese character riddles~\citep{tan2016solving} add wordplay. They rarely require cultural knowledge. Cultural riddle benchmarks exist only for a few languages or locales. In MultiNRC~\citep{fabbri2025multinrc}, 258 out of 1,055 reasoning questions are wordplay and riddles; no model exceeds 50\% on LLM judge scores. BanglaRiddleEval~\citep{sayeedi2026can} features 1,244 traditional Bangla riddles with LLM-generated explanations and distractors; the best multiple-choice accuracy is about 56\%, against 83\% for humans. MonCulture-eval~\citep{minggad-etal-2026-monculture} includes a Mongolian riddle task across two scripts and three regions, scored by LLM judges with a human audit of 10\% of the scores.
\citet{boczka2026hungarianriddles} introduced riddle writing focused on culture-specific associations and clues across a list of selected topics, with an evaluation scheme focused on hallucinations and abstention.
Each of the existing cultural riddle benchmarks covers one to three languages with one locale per language, relies on multiple choice, unvalidated LLM judges, or human scoring alone, and covers few topics and model families. \dataset scales the one-language pilot of \citet{boczka2026hungarianriddles} to \numcommunities communities, with human and validated LLM-judge evaluation.

\section{\dataset Benchmark}

\dataset contains \numriddles original riddles written by \numwriters native speakers for \numcommunities national-linguistic communities. The target is 100 riddles per community: ten for each of eight shared topics, which enable cross-cultural comparison, and ten for each of two topics chosen by the community\footnote{The \numriddles released riddles exclude eight removed for harmful or discriminatory content and one excluded from human evaluation for using wrong language.}.

\subsection{Participatory Benchmark Construction}\label{sec:benchmark-construction}

\paragraph{Community Unit \& Recruiting}
Neither language nor country alone captures cultural context, so our working unit is the national-linguistic community as it exists today (e.g., Hungarian in Hungary, Spanish in Mexico), an imperfect~\citep{alkhamissi2026hire} but scalable basis for data collection and coverage reporting. We recruited volunteer writers through social media and personal networks, requiring only self-identification with a community, proficiency in its language, and basic English. Over six months, \numwriters writers contributed, typically one or two per community (\Cref{app:coverage}). Consent to publish the data was obtained from the writers, and all participants are offered opportunities to contribute to research on top of the collected data.

\paragraph{Task Format and Example}
Each riddle instance contains a topic, an indirectly phrased riddle, and a short reference answer in the local language, with optional hints and comments (\Cref{tab:examples}). Most instances also include English translations, which serve only as reading aids: they may lose information needed to solve the riddle and some concepts simply have no English equivalent. 

\paragraph{Riddle Writing Process \& Instructions}
Each team wrote its riddles from scratch within the shared topic framework, without translating content from other communities. Writers were asked for tricky, indirect questions with a single exact answer (typically 1--4 words), using synonyms, indirect descriptions, local jargon, and nicknames recognizable within their community, so that solving a riddle takes more than a single lookup or web search. In a post-writing survey (\numsurveyrespondents respondents from \numsurveycommunities communities), 97.4\% of respondents reported having drawn on personal knowledge and experience, 68.4\% also on online sources, and 55.3\% on other community members. To calibrate difficulty, 46.1\% tested riddles on AI tools (without revealing the answers); 39.5\% used AI to brainstorm and 5.3\% to paraphrase or co-write.

\paragraph{Quality \& Diversity Assurance}
At least one native speaker other than the author verifies every riddle: it must be solvable, its clues unambiguous, and its reference answer correct. Not every community member must be able to solve it, since some riddles rely on knowledge specific to a generation, region, or interest~\citep{boczka2026hungarianriddles}, and agreement is not a requirement for cultural ``truth''~\citep{alkhamissi2026hire}.
In our survey, one third of riddle writers estimated their riddles would mostly rely on only well-known knowledge, another third also included some regional knowledge, 26\% mostly regional or local knowledge, and only 3\% estimated they would require niche knowledge.

\subsection{Contents and Coverage}\label{sec:contents}

\begin{figure}[t]
    \centering
    \includegraphics[width=\linewidth,trim={0 0 0 5cm}, clip]{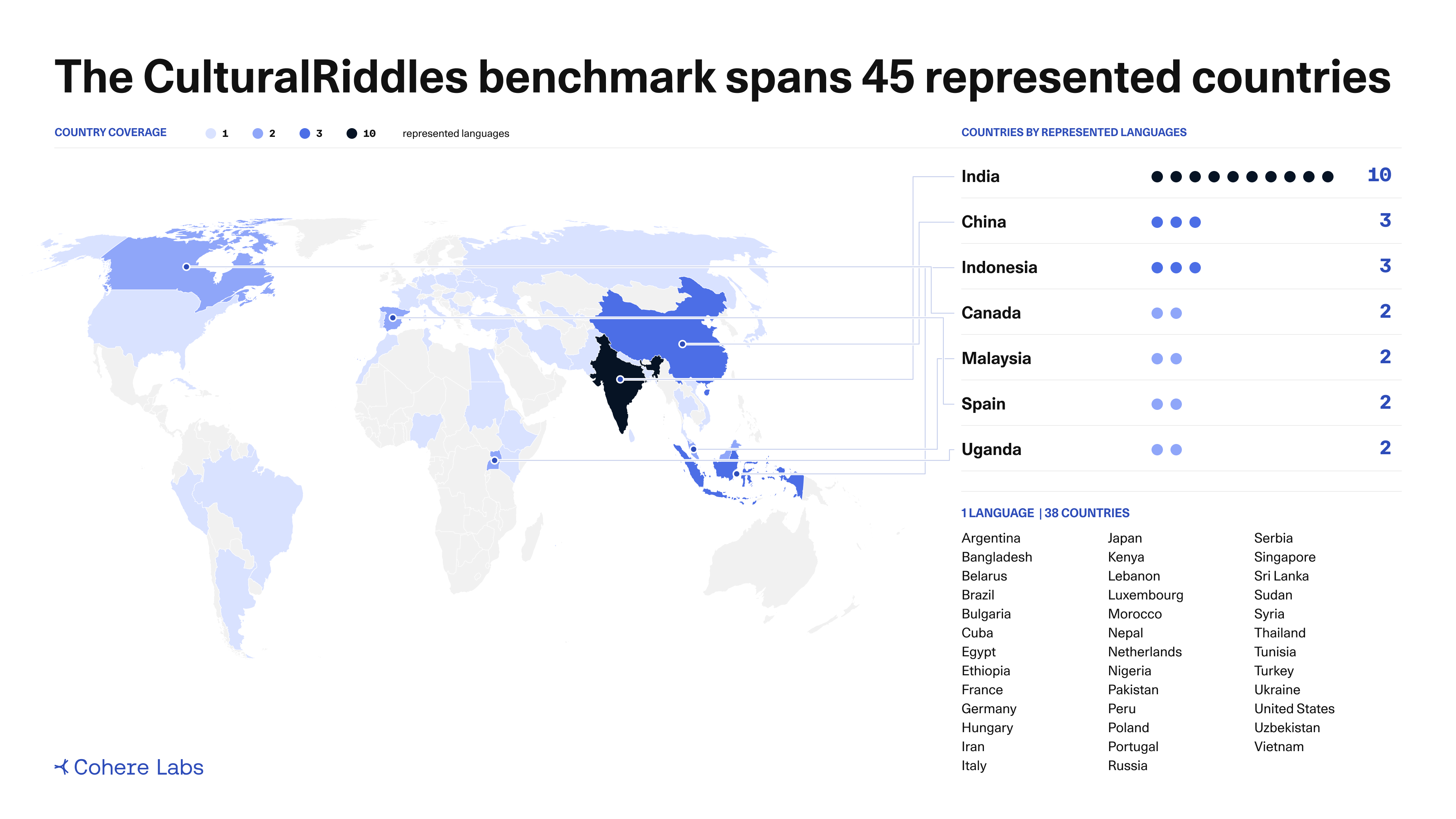}
    \caption{
    \textbf{Cultural riddles across languages and borders.}
    \dataset brings together \numriddles original riddles from \numcommunities communities across \numlang languages and \numcountries countries, including many underrepresented languages and locales. 
    See \Cref{app:coverage} for the full list.}
    \label{fig:map}
\end{figure}

\paragraph{Linguistic and Geographic Coverage}
\dataset (\Cref{fig:map}) includes languages absent from prior cultural benchmarks (e.g., \emph{Buginese}, \emph{Mongondow}, \emph{Masaaba}, \emph{Konkani}, \emph{Sudanese Arabic}) and underrepresented locales (e.g., \emph{Uganda}, \emph{Luxembourg}, \emph{Cuba}). Its language--country mapping is denser than in prior benchmarks (\Cref{tab:benchmark_comparison}): seven languages appear in several countries (e.g., Spanish in Argentina, Cuba, Peru, and Spain), seven countries contribute several languages (e.g., ten for India), and Arabic appears in five varieties.

\paragraph{Topics}
Building on the Hungarian riddles of \citet{boczka2026hungarianriddles}, we selected eight shared topics expected to be relevant across cultures (\emph{Politics, History, Geography, Arts, Sports, National Heritage, Kids' World,} and \emph{Everyday Life}). Additionally, each community could freely choose two ``wildcard'' topics, designated to capture particularly relevant concepts of their culture.
Among the 124 topics chosen by communities, the most common theme is \emph{Food} (27.4\%), followed by \emph{Nature} (15.3\%), \emph{Language} (13.7\%), and \emph{Media} (12.9\%). We also received culture-exclusive topics, such as \emph{Public bath etiquette} by Japanese (Japan) and \emph{Wedding Micro-traditions} by Sindhi (India) communities. Writers rated \emph{Politics} the hardest shared topic to write (3.41/5) and \emph{Everyday Life} the easiest (2.37/5).

\begin{figure}[t]
    \centering
    \includegraphics[width=0.7\linewidth]{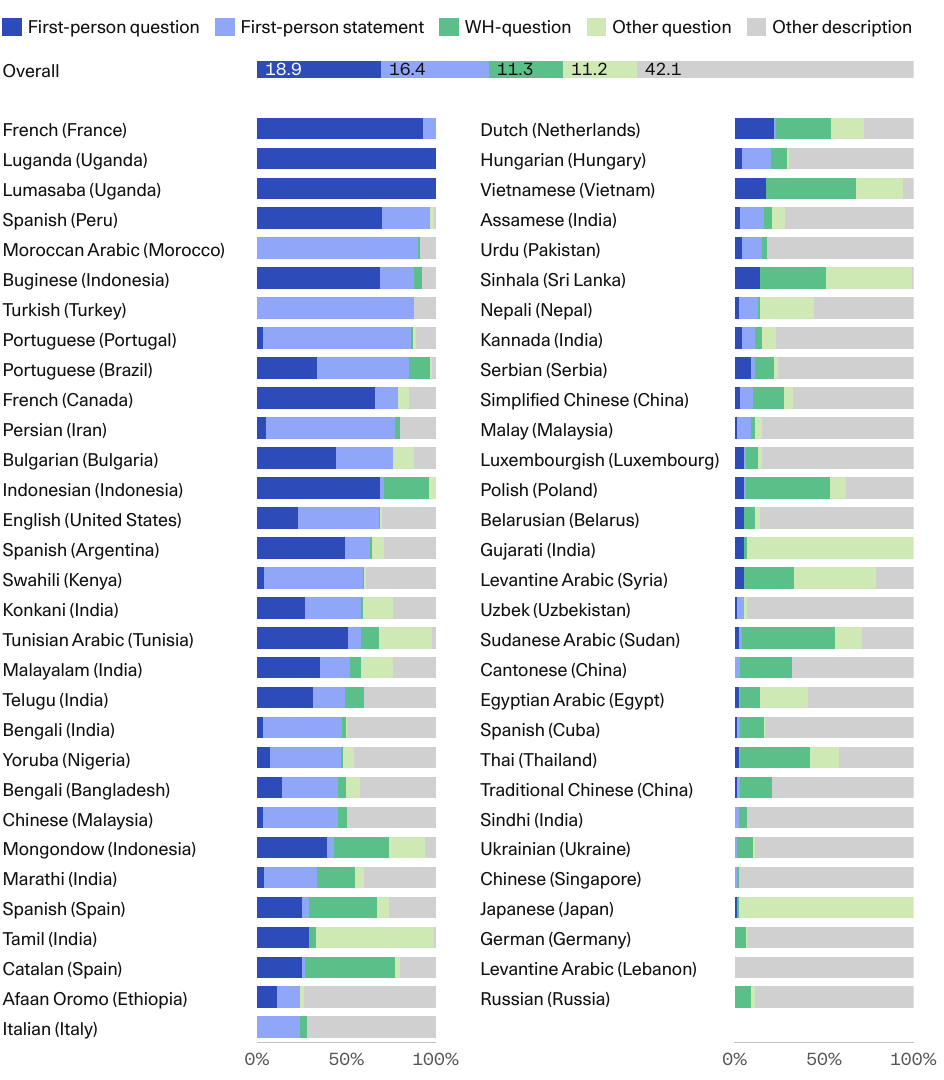}
    \caption{\textbf{Communities favor different riddle forms.} Share of riddles in each form, overall (6,092 English translations, 61 communities) and in one community dominated by each form, labeled by ISO 639-3 language and ISO 3166 country code (\Cref{tab:coverage}). Questions end in ``?'' or open with a question word (WH-questions open with one); first-person riddles contain a first-person pronoun.}
    \label{fig:surface-forms}
\end{figure}

\paragraph{Riddle Patterns \& Content} We imposed no fixed format on the riddles to leave room for unique cultural preferences. The examples in \Cref{tab:examples} illustrate the diversity of riddle formats chosen, e.g., a brief statement in the German and Luxembourgish examples or a first-person description in the Sudanese Arabic example. Many communities favor one form: \formlugandawhoamipercent of Luganda riddles close with a first-person question (``Who am I?''), while \formjapanesewhatisitpercent of Japanese riddles close with a short question of the type ``What is it?'' (\Cref{fig:surface-forms}).
Some riddles also build on the culture's own riddle traditions, e.g., Tamil riddles about \emph{Geography} grounded in old Tamil poetry about landscapes, or Persian riddles built on wordplay, poetic imagery, and rhyme.
Many others draw on collective memory, including painful histories (e.g., betrayal at Plassey (Bangladesh), persecution under Brazil’s dictatorship, imprisonment and survival in Syria, or the contradiction between slavery and ideals of equality in the US). These clues reach beyond language into shared ways of remembering: particularly in \emph{Politics} and \emph{History}, a few words can evoke a larger narrative of loss, injustice, or resistance. Solving them requires familiarity with how an event is remembered, not only with the event itself.

\paragraph{Data and code.}
\dataset is available on Hugging Face under CC-BY-4.0, with the request that it not be used to train models or to generate synthetic data. Code for evaluation and analysis is available at \url{https://github.com/Cohere-Labs-Community/CulturalRiddles}. New communities can contribute through the repository, which provides a riddle template, an automatic format check, and a review process. 

\section{Experimental Setup}
\subsection{Models}\label{sec:setup-models}
We evaluate \nummodels proprietary, open-weight, and open-source model configurations (\Cref{fig:model-summary})\footnote{A model is \emph{open-weight} if it releases weights, code, and a technical report, and \emph{open-source} if it additionally releases its training data and full training recipes.}. They are dense and mixture-of-experts models with disclosed sizes from 0.8B to 1.6T parameters, several reasoning settings, and one search-enabled configuration. Each configuration answers the \numbenchmarkedriddles benchmarked riddles (\numbenchmarkedcommunities communities) once with the same English prompt template including country and topic for context (\Cref{app:benchmarking_details}), yielding \numresponses total responses\footnote{Of the \numriddles released riddles, 100 from Spanish (Spain) arrived after the model runs, and one Spanish (Cuba) riddle was corrected during benchmarking; these 101 are not included in the reported model results.}.

\subsection{Evaluation}\label{sec:evaluation-pipeline}

\paragraph{Metric}
Each answer receives a score of 0 (\emph{incorrect}), 1 (\emph{admits not knowing}), 2 (\emph{partially correct}), or 3 (\emph{correct}).
To capture finer distinctions within these score categories, we define 13 response types, such as \textit{hallucinated} (score~0) and \textit{different valid answer} (score~3), as shown in \Cref{tab:errortypes,tab:annotation_example}.
Our primary metric, \textbf{accuracy}, is the share of scored responses with score~3.
Accuracy differences are reported in percentage points (pp).

\begin{figure}[t]
    \centering
    \includegraphics[width=\linewidth]{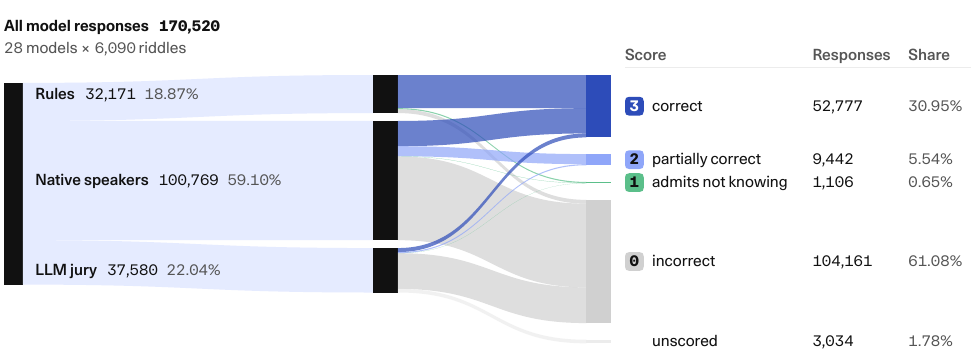}
    \caption{\textbf{Native speakers score more than half of all responses.} Band widths count model responses through the scoring pipeline. The jury assigns a score only when both judges agree; disagreements and missing verdicts remain unscored.}
    \label{fig:pipeline}
\end{figure}

\paragraph{Scoring Pipeline}
Open-ended answers may differ from the reference while remaining correct.
To balance reliability and annotation effort, each of the \numresponses responses takes one of three routes to a score (\Cref{fig:pipeline}): rules, native speakers, or a human-validated LLM jury.

\paragraph{Rule-based Scoring} Rules score case-insensitive exact matches as~3 (\exactmatchpercent of all responses), empty outputs and copies of the prompt's answer placeholder as~0, and explicit ``I don't know'' responses as~1.

\input{tables/human_error_types}

\paragraph{Human Annotation} Community members assign scores (0-3) and error types, recognizing valid local answers beyond the reference answer. Identical answers to the same riddle are labeled once for all models that gave them. Annotators score the ten most frequent unscored answers per riddle and can optionally score more. Their \numuniquehumanresponses unique labels cover \numhumanresponses model responses (\humanevalpercent). Despite thorough instruction (\Cref{app:human_eval}), we have \invalidhumanpercent of responses that are invalid entries, including all Sindhi (India) labels. They are routed to the jury, as with three communities without human labels (\Cref{app:judge_results}). A second native speaker independently relabeled a subset in \numiaacommunities communities, yielding \numcleanediaapairs valid pairs; we call the first label H1 and the second H2.

\begin{table*}[!htbp]
    \centering
    \small
    \begin{tabularx}{\textwidth}{@{}p{0.32\textwidth} c c p{0.2\textwidth} X@{}}
        \toprule
        \textbf{Model answer} & \textbf{Score} & \textbf{Type} & \textbf{Note} & \textbf{Gloss} \\
        \midrule
        Schicksalstag der Deutschen & 3 & 3.1 & & ``Germans' day of fate'' \\
        9. November & 0 & 0.2 & & the date, not its name \\
        der 11. September 2001 & 0 & 0.2 & direct German translation & the U.S.\ attacks \\
        Reichstagsbrand & 0 & 0.3 & unrelated & the 1933 Reichstag fire \\
        der Anschlag vom 11. September 2001 (auch bekannt als 9/11) & 0 & 0.4 & incorrect resolution of date format & U.S.\ attacks, read as 11 September \\
        Ich weiss nicht & 1 & -- & & ``I don't know'' \\
        \bottomrule
    \end{tabularx}
    \caption{\textbf{Annotations for the German riddle} ``Name des deutschen 9/11'' (``Name of the German 9/11''; reference \textit{Schicksalstag} (9 November)) from \Cref{tab:examples}. Six of the 28 model answers are shown with their labels. 
    }
    \label{tab:annotation_example}
\end{table*}

\paragraph{LLM Jury} Two judges, Gemini~3.5 Flash and DeepSeek~V4.1 Flash, score the remaining answers with model names hidden. They receive the same information as annotators and return a score and error type (\Cref{app:judge}). Because judges can be biased~\citep{ye2025justice} and miss cultural nuance~\citep{jin2026juicebenchmarkevaluatingllmjudge}, we develop the prompt on held-out riddles and evaluate three sets of distinct (riddle, answer) pairs:
\textbf{(A)} \numsettingapairs doubly annotated pairs from \numsettingariddles test riddles;
\textbf{(B)} \numsettingbpairs human-labeled test pairs, stratified by community and score across all \numhumanscoredcommunities human-evaluated communities; and
\textbf{(C)} \numsettingcpairs pairs without valid human labels, which permit only judge--judge comparisons.
An answer is only considered scored here when both judges agree on its 0--3 score; unresolved responses (\unscoredpercent of all outputs) are excluded from accuracy computation.

\section{Results}\label{sec:results}

We first assess whether the jury agrees with native speakers (\Cref{sec:results-judges}), then compare models and communities using combined rule, human, and jury scores (\Cref{sec:results-leaderboard,sec:results-exact,sec:results-cultures}) over the \numhumanscoredcommunities communities with valid human evaluation. Error-type analyses use only rule-based scores and human labels (\Cref{sec:results-errors}).

\subsection{LLM judge agreement approaches human agreement}\label{sec:results-judges}
\begin{figure}[t]
    \centering
    \includegraphics[width=\linewidth]{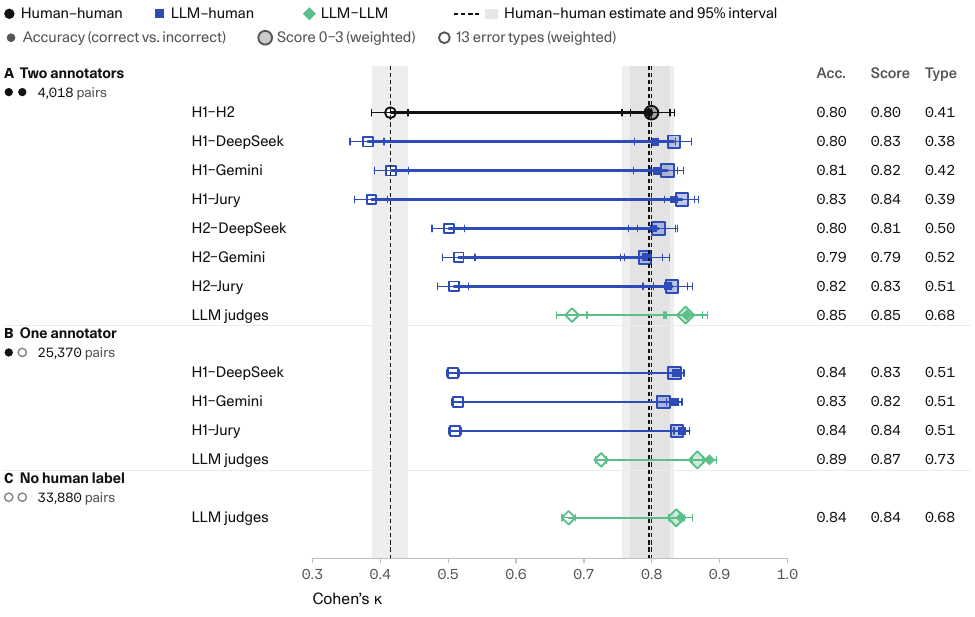}
    \caption{\textbf{Judge agreement approaches human agreement on accuracy and scores, but not on fine-grained types.} Cohen's $\kappa$ for each rater pair in settings A--C (\Cref{sec:evaluation-pipeline}), with 95\% intervals from 1,000 riddle-bootstrap resamples. Solid, pale, and hollow markers denote accuracy, the weighted 0--3 score, and the 13 weighted response types; color and shape identify the rater pair. Shaded bands mark the human--human intervals from setting A, not equivalence tests in B or C. ``Jury'' takes the lower judge score and that judge's type (DeepSeek's on a tie), including disagreements rather than leaving them unscored as in production scoring (\Cref{app:judge}).}
    \label{fig:rater-agreement}
\end{figure}

\paragraph{Human annotators agree on scores but less on error types.}

On the doubly annotated sample (setting~A, \Cref{fig:rater-agreement}), humans agree on accuracy at Cohen's $\kappa=0.80\pm0.04$ and on the 0--3 score at quadratic-weighted $\kappa=0.80\pm0.03$~\citep{landis1977measurement}. Accuracy agreement ranges across communities from $\kappa=0.59$ (\textit{Ukrainian}) to 1.00 (\textit{Luganda}), excluding two communities with too few correct answers. Agreement on the 13 error types is weaker ($\kappa=0.41$ with partial credit within a score, 0.31 without), mostly because the guideline left room for subjective interpretation: \textit{hallucinated} versus \textit{right category, wrong entity}, when to award \textit{partial credit}, and whether a \textit{different valid answer} needs its own label rather than \textit{correct}. We therefore use error types descriptively (\Cref{app:human_eval}).

\paragraph{LLM--human agreement is comparable to human--human agreement.} In setting~A, individual zero-shot judges agree with H1 and H2 at $\kappa=0.79$--$0.81$ on accuracy and quadratic-weighted $\kappa=0.79$--$0.83$ on the 0--3 score, close to human--human agreement on both. Agreement with H1 remains high in setting~B: $0.83$--$0.84$ on accuracy and $0.82$--$0.83$ on scores. Ten in-prompt examples change accuracy $\kappa$ by about 0.01 or less, so we use zero-shot judges. Like humans, judges agree less on fine-grained response types ($0.38$--$0.52$).

\paragraph{Human disagreement is more frequent when judges disagree.}
The two LLM judges agree highly on accuracy in all three settings (\Cref{fig:rater-agreement}). In setting~A, humans disagree on accuracy for 27\% of pairs where the judges disagree, 8.4 times the rate where they agree, and on the full 0--3 score for 39\%, 5.1 times the rate where they agree. Judge disagreement thus flags answers that are also difficult for humans to score.



\paragraph{Judges share systematic errors.}
The judges share about two-thirds of their errors. First, they are \emph{lenient on the strongest models}: they jointly accept 9.0\% and 6.9\% of human-rejected answers from Gemini~3.1 Pro and GPT-5.5, respectively (\Cref{fig:judge-by-model}). Second, they \emph{reject valid alternatives}: individual judges reject 25--29\% of \textit{different valid answers}, versus about 5\% of answers labeled simply \textit{correct}. Third, \emph{agreement varies across communities}, ranging from $\kappa=\jurykappamin$ for \textit{\juryminlanguage} to \jurykappamax for \textit{\jurymaxlanguage} (\Cref{fig:judge-by-community}).


\subsection{Frontier models perform well, but most models lag behind}
\label{sec:results-leaderboard}

\begin{figure}[t]
    \centering
    \includegraphics[width=\linewidth]{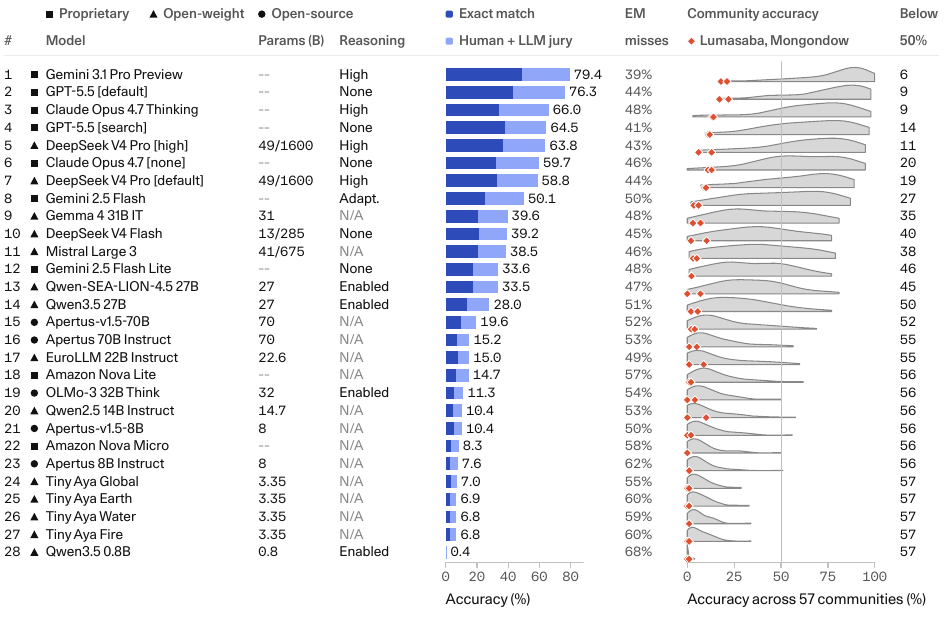}
    \caption{\textbf{High overall accuracy can coexist with cultural blind spots.} Stacked bars pool scored responses: dark blue is exact-match accuracy and light blue adds answers accepted by native speakers or the jury; \emph{EM misses} is the share of ultimately correct responses missed by exact match. Half-violins weight each of the \numhumanscoredcommunities community accuracies equally, orange diamonds mark Lumasaba and Mongondow, and the $<50\%$ column counts communities below that threshold. Parameter counts are in billions (active/total for mixture-of-experts); access categories follow \Cref{sec:setup-models}.}
    \label{fig:model-summary}
\end{figure}

\paragraph{Larger models generally perform better, but not always}
Gemini~3.1 Pro Preview leads at \bestmodelaccuracy, followed by GPT-5.5 (76.3\%) and Claude Opus~4.7 Thinking (66.0\%; \Cref{fig:model-summary}). The best open model, DeepSeek~V4 Pro with high reasoning, ranks fifth at 63.8\%. Apart from other configurations of these models and Gemini~2.5 Flash (50.1\%), every model scores below 40\%, and models of 8B parameters or fewer reach at most 10.4\%. Yet size alone does not explain performance: Gemma~4 31B (39.6\%) outperforms both the similarly sized OLMo~3 32B Think (11.3\%) and the larger Apertus 70B models (15.2\% and 19.6\%).

\paragraph{Regional models show no extra advantage in their target regions.}
We test whether regionally adapted models gain more in their target regions than elsewhere. Regional Tiny Aya variants do not differ significantly within any region (Cochran's $Q$, all $p>0.48$). Qwen-SEA-LION-4.5 27B gains 5.5\,pp over Qwen3.5 27B overall, but we find no significant extra gain in its target regions.

\paragraph{Reasoning helps, but web search can hurt.}
Thinking raises Claude Opus~4.7's accuracy by 6.3\,pp, and high reasoning effort adds 5.0\,pp for DeepSeek~V4 Pro. Web search, by contrast, lowers GPT-5.5's accuracy by 11.8\,pp. Among answers that GPT-5.5 gives correctly without search but incorrectly with it and that humans assigned an error type, 39\% name the wrong entity in the right category and 25\% are hallucinated. Search may reinforce familiar keyword associations: a Swahili (Kenya) riddle asks for the ``home of the Lunatic Express'', but multiple models return the Uganda Railway itself. They identify the entity associated with ``Lunatic Express'' while overlooking ``home'', which points to the intended Railway Museum in Nairobi.

\begin{figure}[t]
    \centering
    \includegraphics[width=\linewidth]{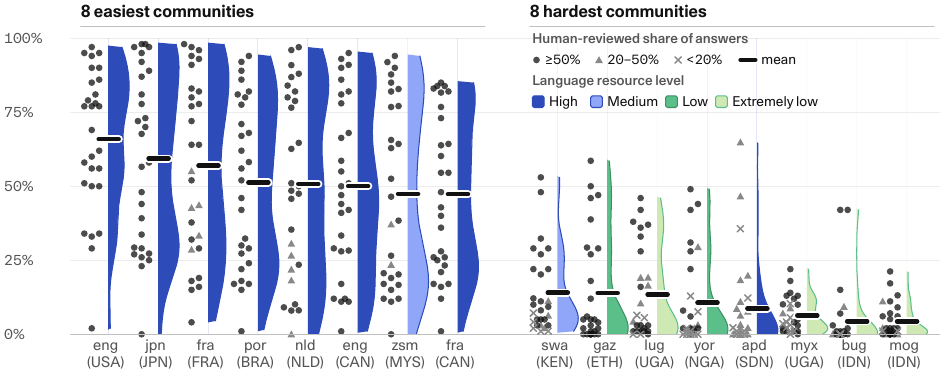}
    \caption{\textbf{Even the strongest models struggle in the hardest communities.}
    Communities are ranked by mean accuracy, weighting models equally; each raincloud shows model accuracies, their mean (black bar), and a density scaled to equal peak width. Color indicates language-resource levels defined by Common Crawl page counts (\Cref{fig:resourcedness}); marker shapes indicate native-speaker review coverage among responses for each model routed beyond automatic rules (\Cref{app:markers}).}
    \label{fig:cultures}
\end{figure}

\subsection{Exact match misses half of correct answers}\label{sec:results-exact}

Exact match credits only responses matching the reference answer (case-insensitive), missing about 47\% of responses ultimately scored correct across the human-evaluated communities (\Cref{fig:model-summary}). Among the \numerrortyperesponses responses scored by rules or humans, native speakers judged an additional \correctvariantpercent correct without a subtype recorded and labeled \differentvalidpercent as \textit{different valid answers} (\Cref{tab:errortypes}).

Human and jury review raises accuracy by 64--166\% relative to exact match for every model except Qwen3.5 0.8B, whose larger relative gain (213\%) reflects a very small exact-match count. Gains generally grow toward weaker models, so exact match can exaggerate performance gaps and change rankings; for example, Claude Opus~4.7 Thinking moves from fifth to third after review.

\subsection{Models hallucinate rather than abstain}\label{sec:results-errors}

The most frequent error across riddles is a \textit{hallucinated} answer, one that is made up or unrelated to the riddle (\hallucinationpercent of the \numerrortyperesponses responses scored by rules or humans). In contrast, only \declinedpercent of these responses admit not knowing, even though the prompt asks models to answer ``I don't know'' when they do not know (\Cref{fig:benchmark_prompt}). Hallucinations shrink as accuracy rises (\Cref{fig:error_breakdown} in \Cref{app:error_types}), but abstention stays rare for every model (at most \declinedmaxpercent for any model).

\subsection{High averages hide cultural blind spots}\label{sec:results-cultures}

\begin{figure}[t]
    \centering
    \includegraphics[width=0.55\linewidth]{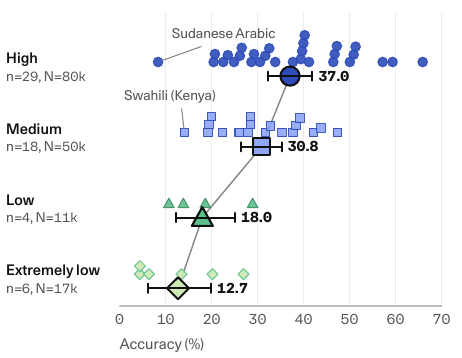}
    \caption{\textbf{Language resources do not explain all community differences.} Small markers show community accuracies pooled over models; large diamonds pool responses within resource levels, with 95\% intervals from 10,000 community-bootstrap resamples. Levels use Common Crawl page counts (CC-MAIN-2026-34): high $\geq$10M, medium 100K--10M, low 10K--100K, extremely low $<$10K or not detected; dialects inherit their macro-language's count. $n$: communities; $N$: scored responses.}
    \label{fig:resourcedness}
\end{figure}

\paragraph{Most models agree on what is easy.}
The eight easiest communities---English (USA), Japanese, French (France), Portuguese (Brazil), Dutch, English (Canada), Malay, and French (Canada)---have mean accuracies of 47--66\% across models, and seven of them use high-resource languages (\Cref{fig:cultures}), defined by Common Crawl page counts (\Cref{fig:resourcedness}). At least 20 of \nummodels models exceed 20\% accuracy in each, and the median model reaches 73\% on English (USA), plausibly because these cultures are well documented on the web and hence in LLM training data.

\paragraph{African and Indonesian communities challenge every model.}
The eight hardest communities are Swahili (Kenya), Afaan Oromo, Luganda, Yoruba, Sudanese Arabic, Lumasaba, Buginese, and Mongondow (mean accuracy 4--14\%; \Cref{fig:cultures}). Here 15--25 of \nummodels models score below 10\%, and the median model scores only 0--9\%. Even Gemini~3.1 Pro and GPT-5.5 average just 43\% and 39\% across these eight, about half their overall accuracy. No model exceeds 22\% on Lumasaba or Mongondow.

\paragraph{Low-resource languages increase the challenge.}
Six of the eight hardest communities use low- or extremely low-resource languages (fewer than 100K pages in Common Crawl). Across all human-evaluated communities, pooled accuracy falls from \poolaccuracyhigh for high-resource languages to \poolaccuracymedium for medium, \poolaccuracylow for low, and \poolaccuracyextremelylow for extremely low-resource languages (\Cref{fig:resourcedness}), yet community accuracies still vary with a standard deviation of \resourcesdmin--\resourcesdmax\,pp within each level. Swahili is medium-resource, and Sudanese Arabic inherits the high-resource classification of Arabic: neither escapes the hardest group.

\section{Analysis}
We first examine what makes a riddle difficult and how language contributes (\Cref{sec:discussion-difficulty}), then pose the riddles as crosswords (\Cref{sec:discussion-formats}), and finally estimate the effort of evaluating new models (\Cref{sec:discussion-new-models}).

\subsection{What makes a riddle difficult?}\label{sec:discussion-difficulty}

\begin{figure}[t]
    \centering
    \includegraphics[width=0.6\linewidth]{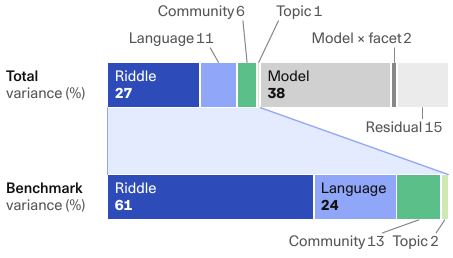}
    \caption{\textbf{Most riddle difficulty is specific to the individual riddle.} Variance components of a crossed random-effects logistic model on the latent log-odds scale (residual variance $\pi^2/3$). The lower bar rescales only the riddle, language, community, and topic components to 100\%; colors and hatching match across bars.}
    \label{fig:difficulty-variance}
\end{figure}

\paragraph{Difficulty mostly comes from the riddle, then the language.}
To separate model ability from riddle difficulty, we fit a crossed random-effects logistic model~\citep{bates2015lme4} to 129,928 scored responses from model--community pairs with at most 2\% unscored responses.
Here, variance measures the spread in the estimated likelihood of a correct answer on the model's latent log-odds scale.
Model identity accounts for 38\% of the total variance (\Cref{fig:difficulty-variance}).
Among the components associated with the benchmark rather than the answering model, individual riddles account for about 60\%, language 24\%, community within a language 13\%, and topic 2\%.
Models largely agree on which languages and communities are hard: interactions between model identity and benchmark factors account for less than 2\% of total variance.

Riddle features explain only 3.1\% of the riddle-level variance (\Cref{app:riddle-features}), but rare answer terms matter.
About a third of the riddles contain a very rare word in their English reference answer, often a local term such as \emph{bebinca}, a Goan layered pudding (rarity measures English word frequency, not cultural knowledge directly), and their adjusted odds of a correct answer are roughly half those of riddles with common answers (OR 0.54, 95\% CI 0.46--0.65).
For example, the Konkani riddle ``Layers of sweet, made during Christmas, what is this?'' asks for \emph{bebinca}, but GPT-5.5 and Claude Opus~4.7 Thinking answer \emph{kuswar}, the platter of Goan Christmas sweets that includes it.
They identify the dish family but miss the specific dish.

\paragraph{Language coverage is not community coverage.}
Language resources remain associated with accuracy after accounting for model identity and riddle features, consistent with the resource disparities documented by \citet{joshi-etal-2020-state} and the cross-language gaps in INCLUDE~\citep{romanou2025include}.
In our feature model, the adjusted odds of a correct answer for low- and extremely low-resource languages are 0.10 (95\% CI 0.02--0.48) and 0.04 (0.01--0.15) times those for high-resource languages, so a high-resource riddle has roughly 10 and 25 times the odds.
The medium-resource estimate (0.49, 0.20--1.21) is less certain (\Cref{app:riddle-features}).

Even among communities sharing both a language and a script, accuracy differs by 5--26\,pp (\Cref{app:riddle-features}).
For example, Simplified Chinese (China) averages 28.5\% versus 46.3\% for Chinese (Singapore), a gap of 17.8\,pp.
Language-level averages thus conceal differences between the communities that a multilingual model is meant to serve.

\paragraph{Translating into English helps, but does not close the gap.}

To estimate how much of this difficulty comes from language alone, we benchmark five models (GPT-5.5, Gemini~3.1 Pro, Claude Opus~4.7, DeepSeek~V4 Pro, Qwen-SEA-LION-4.5 27B) on English translations of the riddles from five hardest communities (\Cref{sec:results-cultures}).
After excluding \numenglishrevealing translations that contain the full reference answer as a clue (e.g., ``I am the art of making pottery'' for \emph{pottery}; 38 from Lumasaba), \numenglishkept riddles remain; English responses are scored by rules and the LLM jury without native-speaker review (\Cref{app:english}).
On the 2,049 responses scored in both conditions, accuracy rises from \englishoriginalaccuracy to \englishaccuracy (paired gain +\englishgain\,pp, 95\% CI \englishgainlow--\englishgainhigh; \Cref{fig:english}).
Buginese, Lumasaba, and Yoruba gain 14--18\,pp, while Sudanese Arabic shows no detectable gain. The same models reach about \englishhighresourceaccuracy on high-resource communities, so translation closes only about a fifth of the gap: much of the difficulty persists in English. 

\subsection{Beyond open-ended answers: Crosswords}\label{sec:discussion-formats}
Because each riddle has a short, exact answer, \dataset can also be posed in other formats. We pose the riddles as crosswords, a long-standing testbed for language models~\citep{wallace-etal-2022-automated,saha-etal-2025-language,leng2025crosswordbench}, which adds checkable constraints: each answer must fit its slot and share letters with the answers that cross it. We built 53 text grids from the communities' riddle submissions (5,057 entries)\footnote{Chinese varieties, Thai, Kannada, and Buginese were excluded because their grids could not be built reliably.} and gave the same clues to 11 model configurations in five matched conditions: plain batch, batch with answer lengths, lengths plus a quarter of each answer's letters, crossword with list output, and crossword with grid output.

We compare crossword input with the length condition, so that both see the same lengths (\Cref{app:crossword}). Crossing structure changes reference agreement by only $+1.6$ points (95\% CI $-0.3$ to $3.7$), but raises the share of consistent crossings from 53.0\% to 60.1\% ($+7.2$, 4.8--9.6): the answers agree more with each other without matching the references more often. Filling the grid is a separate skill that varies widely between models: Gemini~3.7 Flash scores 75.6\% with list output but 5.4\% with grid output, Claude Opus~4.8 returns no usable grid for any puzzle, and GPT-6 Astra places 88.3\% of answers correctly.


\subsection{Effort of benchmarking new models}\label{sec:discussion-new-models}
We recommend scoring a new model on \dataset in four steps, each passing fewer answers onward: (1) \textbf{rules} for exact matches, empty outputs, placeholders, and explicit admissions of not knowing (\autoresolvedpercent of our responses); (2) \textbf{existing human labels} for answers already given to the same riddle; (3) the \textbf{LLM jury}, which keeps a score only when both judges agree; and (4) \textbf{native speakers} for unresolved answers, with extra review in communities where agreement with human labels is low (\juryuncertaincommunities;\Cref{app:judge}).
Treating each model as new and using only rules and human labels from the other models (\Cref{app:new-model-effort}), these steps already cover about 80\% of the leading models' responses, since strong models tend to repeat answers that other models gave. Coverage falls to 51\% for the median model and 2--13\% for the smallest models, whose answers are mostly unique. Roughly 100--115 answers per leading model would still need native review.





\section{Discussion \& Conclusion}
We built \dataset, a participatory benchmark of \numriddles original riddles from \numcommunities communities (\numcountries countries, \numlang languages), to test how LLMs combine cultural knowledge with reasoning. Native speakers write, verify, and score its riddles, recognizing answers that exact match misses and judges sometimes reject. Frontier models lead overall, yet the top three proprietary configurations each score below 50\% in 6--9 of the \numhumanscoredcommunities human-evaluated communities, and the best open configuration does so in 11. No model exceeds 22\% on Lumasaba or Mongondow. Models rarely admit not knowing. Evaluations that treat abstentions and wrong answers alike can reward guessing~\citep{kalai2025why,wen-etal-2025-know}; our 0--3 scale tracks abstention alongside accuracy.

Translation closes only a fifth of the hardest communities' gap to high-resource communities, communities sharing a language and script differ by up to 26\,pp, and regional adaptation shows no detectable extra advantage in its target regions. Stronger multilingual ability alone is therefore unlikely to close these gaps; much of what models lack appears to be content from the communities themselves.

The largest difficulty lies in what the riddles are built on: associations, traditions, perceptions, and memories shared within a community. Community members write and verify these riddles, so this knowledge is available to them by construction, yet the best models still miss it in many communities; final-answer scores cannot show whether they lack facts or misread clues. Because scoring needs native-speaker review for only about 100 answers per leading model (\Cref{sec:discussion-new-models}), \dataset can grow to new models and communities and track how far future models move this cultural frontier.

\section{Ethics Statement}
Riddle writers and annotators are members of the communities they represent and contributed as unpaid volunteers. All contributors joined under a project proposal that described \dataset as an open benchmark to be released publicly; a follow-up form asked the authors how they wish to be credited. The released dataset credits writers at the community level and does not attribute individual riddles. We report the writer survey only in aggregate. Some riddles refer to painful histories, such as persecution, dictatorship, or slavery. We keep them because they are part of how these communities remember events. After a content review, we removed 8 riddles and added a content note to 56 others; the dataset card includes a content disclaimer. The data released contains no names or contact details of contributors.  

\section{Acknowledgments \& Disclosure of Funding}
We thank Thomas Euyang for the help with the visualizations, and Brittawnya Prince for the coordination of the Cohere Labs Community Discord. 
We thank Catherine Arnett for her wise advice on embarking on a massive collaborative benchmarking journey.
We thank the reviewers of the MRL workshop for their helpful feedback to an earlier draft.

Arora’s lab is partly supported by grants from the Novo Nordisk Foundation (NNF24OC0099109), the Pioneer Centre for AI, EU Horizon 2020 (101168951), and Danish Advanced Research Academy (DARA) and Danish Data Science Academy (DDSA) PhD Fellowships. We also gratefully acknowledge generous gifts from Microsoft and IT-vest - networking universities.

\clearpage
\bibliography{custom}

\appendix

\section{Reproducibility Details}\label[appendix]{app:repro}
\renewcommand{\dbltopfraction}{0.95}
\renewcommand{\dblfloatpagefraction}{0.8}

\subsection{Benchmark Coverage}\label[appendix]{app:coverage}
\Cref{tab:coverage} lists the communities in \dataset, each identified by a language and a country.

\input{tables/coverage}

\subsection{Benchmarking Prompt and Models}\label[appendix]{app:benchmarking_details}
All models receive the prompt in \Cref{fig:benchmark_prompt}. Open-weight and open-source models range from 0.8B to 1.6T parameters and include general multilingual and reasoning families (Qwen~\citep{team2026qwen3}, Gemma~\citep{team2025gemma}, Mistral~\citep{jiang2023mistral}, OLMo~\citep{olmo2025olmo3}) and models built for multilingual or regional coverage (Tiny Aya~\citep{salamanca2026tiny}, EuroLLM~\citep{ramos2026eurollm22b}, Apertus~\citep{hernandezcano2025apertus, apertus2026}, SEA-LION~\citep{ng2025sealion}). Proprietary models include GPT-5.5~\citep{openai2026gpt55}, Claude Opus~4.7~\citep{anthropic2026opus47}, and Gemini~3.1~\citep{google2026gemini31}, with their reasoning and web-search variants (\Cref{fig:model-summary}).

\begin{figure}[!htbp]
\centering
\begin{tcolorbox}[
    enhanced,
    width=\linewidth,
    colback=tzBlueFill,
    colframe=tzBlueBorder,
    boxrule=1.2pt,
    arc=6pt,
    left=5pt,
    right=5pt,
    top=4pt,
    bottom=2pt,
    title={\small Benchmarking prompt},
    coltitle=white,
    colbacktitle=tzBlueHeader2,
    fonttitle=\bfseries,
  ]
  \scriptsize
  \begin{minipage}[t]{0.48\linewidth}
  You are an AI assistant; your goal is to solve riddles in a culturally specific
  riddle benchmark.

  \smallskip
  You are solving a cultural riddle from
  \texttt{\{community\_display\}}. The riddle is written in
  \texttt{\{community\_display\}} (\texttt{\{script\_name\}}), and its topic is
  \texttt{\{topic\}}.

  \smallskip
  Use this cultural and linguistic context to interpret the
  riddle, reason about its clues, and provide your final answer.

  \medskip
  \textbf{Input Format}

  \smallskip
  The riddle is provided as plain text on the line(s) following a
  \texttt{RIDDLE:} marker:

  \smallskip
  \texttt{RIDDLE:}\\
  \texttt{<the riddle text>}

  \medskip
  \textbf{Input Example}

  \smallskip
  \texttt{RIDDLE:}\\
  \texttt{I speak without a mouth and hear without ears. I have no body, but I
  come alive with wind. What am I?}

  \medskip
  \textbf{Instructions}
  \begin{itemize}[leftmargin=1.5em, itemsep=0.1pt, topsep=2pt]
    \item Read the riddle carefully.
    \item Use both the content of the riddle and any relevant cultural knowledge
      associated with the specified community.
    \item Provide the answer you believe is most likely correct, and write your
      answer in \texttt{\{community\_display\}} (\texttt{\{script\_name\}}).
    \item If you do not know the answer, return \texttt{"I don't know"} as the
      \texttt{"riddle\_answer"}. Do not invent an answer simply to be responsive.
    \item Your reasoning should be brief (1--2 sentences), explaining the key
      clues you used and why you selected the answer. If uncertain, briefly state
      what information is missing or why the riddle is ambiguous.
  \end{itemize}

  \end{minipage}\hfill
  \begin{minipage}[t]{0.48\linewidth}
  \textbf{Output Requirements}
  \begin{itemize}[leftmargin=1.5em, itemsep=0.1pt, topsep=2pt]
    \item Return ONLY a valid JSON object.
    \item NEVER output any text before or after the JSON object.
    \item Do NOT use double quotes (\texttt{"}) inside the
      \texttt{"riddle\_answer"} or \texttt{"reasoning"} values, as they will
      break the JSON. If you need to quote something, use single quotes
      (\texttt{'}) instead.
  \end{itemize}

  \medskip
  \textbf{Output Template}

  \smallskip
  \begin{jsonbox}
    \ttfamily\scriptsize
    \{\\
    \hspace*{1.2em}"riddle\_answer": "<your\_riddle\_answer>",\\
    \hspace*{1.2em}"reasoning":\\
    \hspace*{2.4em}"<your\_reasoning\_in\_1\_2\_sentences>"\\
    \}
  \end{jsonbox}

  \medskip
  \textbf{Output Example}

  \smallskip
  \begin{jsonbox}
    \ttfamily\scriptsize
    \{\\
    \hspace*{1.2em}"riddle\_answer": "an echo",\\
    \hspace*{1.2em}"reasoning": "The riddle describes something that speaks without a
    mouth and is heard without ears, which points to an echo."\\
    \}
  \end{jsonbox}
  \end{minipage}
\end{tcolorbox}
\caption{\textbf{The benchmarking prompt} used for all models. Placeholders in braces are filled per riddle with its specific context, e.g. ``Konkani, India'' for \texttt{community\_display} and ``Devanagari script'' for \texttt{script\_name}.}
\label{fig:benchmark_prompt}
\end{figure}

\paragraph{Compute.} The runs used about 217M tokens in total (119M input, 98M output; one batch of 1,700 Gemini~3.1 Pro answers has no token counts). Inputs average 455--730 tokens per riddle (about 1,700 for GPT-5.5 and 4,900 with web search, including retrieved context). Outputs range from about 70 tokens per riddle (Gemma~4, Nova) to 6,700 (OLMo~3 Think). Because runs mixed hosted APIs and locally served models, we report tokens rather than cost.

\paragraph{Limitations of single runs.} Each configuration answered each riddle once with one prompt, so rankings could change under prompt variations~\citep{mizrahi-etal-2024-state}, and scores might not be reproducible under stochastic decoding.

\subsection{Human Evaluation Protocol} \label{app:human_eval}

\paragraph{Procedure.} Each community's spreadsheet lists distinct answers not scored by rules, with the riddle, topic, reference in the original language and English, and model answer, but no model names (\Cref{tab:annotation_example}). Annotators enter a consistent score and type (e.g., score~0 requires a 0.x type), adding notes for ``Other'' types. They can consult community members or the author about unclear references. Validation rejects inconsistent rows and routes their responses to the jury.

\paragraph{Second annotation.} A second native speaker independently relabeled \numcleanediaapairs pairs from \numcleanediaariddles riddles in \numiaacommunities communities. Setting~A retains \numsettingapairs pairs from \numsettingariddles test riddles, excluding 36 placeholder answers scored by rules and 192 pairs from prompt-development riddles. Human correctness agreement ranges from $\kappa=0.59$ (Ukrainian) to 1.00 (Luganda), excluding two communities with too few correct answers for a stable estimate. Volunteer availability limits double annotation; it does not cover every community, including most of the hardest ones.

\paragraph{Guidelines.} Beyond \Cref{tab:errortypes}, the guidelines ask annotators to verify that a \textit{hallucinated} (0.3) entity does not exist in the culture (e.g., a dish assembled from real ingredients), to label an answer from a dominant or neighboring culture as \textit{closely related culture or region} (2.1; e.g., a Karnataka dam for an ancient Tamil one), and to reserve score~1 for explicit admissions of not knowing.

\paragraph{Reproducibility of human annotations.} Our 13 types extend the 0--3 scores of \citet{boczka2026hungarianriddles}. We refined them on 500 responses from five communities. Annotators and judges disagree most on boundaries such as hallucinated versus too broad or right category, wrong entity (\Cref{fig:rater-agreement}). Some disagreements may reflect legitimate differences rather than noise~\citep{davani-etal-2022-dealing}, but clearer definitions could improve reproducibility.

\paragraph{Answers, not reasoning.} We score final answers, not reasoning traces. Correct answers can follow a misreading or a lucky guess; wrong answers can follow a sound interpretation that misses one local fact. Accuracy therefore does not identify which reasoning step fails.

\subsection{LLM Judge Protocol}\label[appendix]{app:judge}
\input{tables/judge_prompt}

\paragraph{Judges and prompt.} Judges receive the riddle, community, topic, reference answer in both languages, candidate answer, and annotation guidelines (\Cref{fig:judge_prompt}). Gemini~3.5 Flash uses structured output with default reasoning; DeepSeek~V4.1 Flash uses a strict forced function call without reasoning. Both use temperature~0 and up to four attempts. We accept only consistent scores and types; all requests in A and B returned valid verdicts.

\paragraph{Development and test split.} A seeded hash assigns 15\% of each community's riddles to development and 85\% to test. Development provides 25 human-labeled tuning answers per type and ten few-shot examples, one per type, from ten further riddles.

\paragraph{Sampling in setting B.} Each community contributes at least 200 human-labeled test pairs (or all if fewer): all score-1 pairs, at least 40 each with scores 2 and 3, and the rest sampled proportional to model-response counts. We enlarge \numsettingbraisedcommunities samples, up to their populations, targeting 95\% CI half-widths near 0.10 for judge--human $\kappa$.

\paragraph{Analysis sample and token use.} Of \numsettingbsampled sampled pairs (effective sample size \settingbeffective), \numsettingbpairs have usable text, valid types, and valid judge verdicts. Input/output token use is \settingbtokensgemini for Gemini~3.5 Flash and \settingbtokensdeepseek for DeepSeek~V4.1 Flash.

\paragraph{Units and weighting.} Setting~A weights distinct pairs equally; B and C weight by model-response counts, with inverse sampling weights in B. The 95\% intervals resample riddles: within communities for B and C, and within the four-rater sample for A (\Cref{fig:rater-agreement}).

\paragraph{Jury summaries.} The jury in \Cref{fig:rater-agreement} takes the lower judge score; official scoring requires identical scores. The hybrid in \Cref{fig:judge-by-model,fig:judge-by-community} instead retains human labels on disagreements, so it still relies on native speakers. Setting~C has no human reference.


\paragraph{Shared judge errors.} Gemini alone accepts 24.6\% and 19.8\% of human-rejected Gemini~3.1 Pro and GPT-5.5 answers, respectively. Individual judges accept about 58\% in the highest reference-similarity band (80--100). Hybrid agreement reaches $\kappa=\jurykappahigh$ for high-resource and \jurykappaextremelylow for extremely low-resource languages.

\paragraph{Native-review criterion.} We recommend extra native-speaker review where the hybrid's setting~B 95\% interval for agreement with human labels extends below $\kappa=0.6$ (\juryuncertaincommunities; \Cref{fig:judge-by-community}), despite retaining human labels on disagreements.

\paragraph{Human-review markers.}\label[appendix]{app:markers} Markers in \Cref{fig:cultures} show human review within each model--community cell, excluding rule-scored responses (exact matches, empty outputs, placeholders, and explicit ``I don't know''). Circles indicate at least 50\% human review, triangles 20--50\%, and crosses below 20\%. These indicate coverage, not confidence.

\clearpage
\section{Additional Results}\label[appendix]{app:results}

\subsection{Error Types by Model}\label[appendix]{app:error_types}
\begin{center}
    \includegraphics[width=\linewidth]{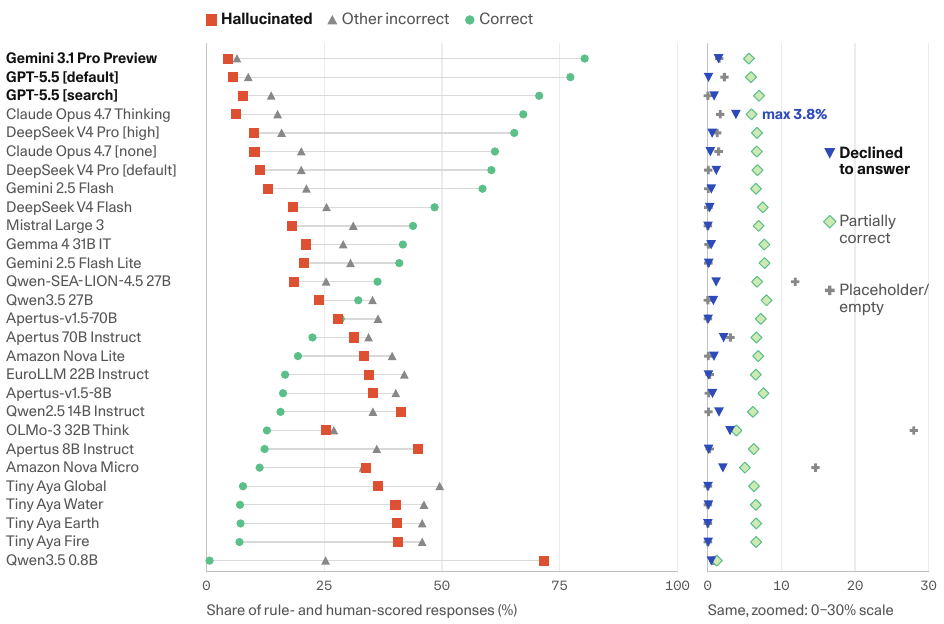}
    \captionof{figure}{\textbf{Hallucinations dominate errors; abstention remains rare.} Shares use only each model's rule- and human-scored responses, excluding jury and unresolved responses; they therefore differ from the accuracy population in \Cref{fig:model-summary}. The panels use different scales, and rows follow exact-match share.}
    \label{fig:error_breakdown}
\end{center}
\vspace{1ex}
\renewcommand{\dbltopfraction}{0.95}
\renewcommand{\dblfloatpagefraction}{0.8}

\subsection{Riddle and Answer Features}
\begin{figure}[t]
    \centering
    \includegraphics[width=\linewidth]{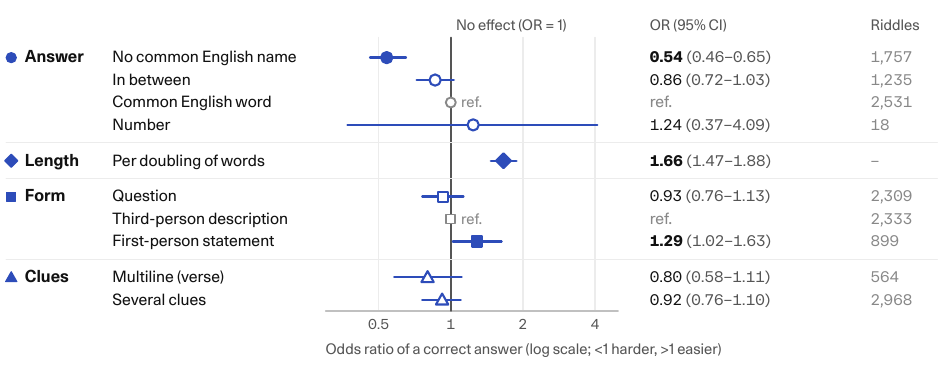}
    \caption{\textbf{Riddle features explain little of the riddle-level variance.} Adjusted odds ratios with 95\% Wald intervals from 125,988 responses to 5,541 riddles; counts refer to distinct riddles. Filled markers exclude 1, hollow markers include 1, and gray markers mark reference levels. The model controls for answering model, resource level, and crossed riddle, language, community, and topic effects.}
    \label{fig:riddle-features}
\end{figure}

\label[appendix]{app:riddle-features}
\Cref{fig:riddle-features} shows the riddle and answer features that \Cref{sec:discussion-difficulty} adds to the random-effects model of a correct answer, and \Cref{fig:surface-forms} shows how riddle forms vary across communities.

The variance model covers 5,688 riddles from all \numhumanscoredcommunities human-evaluated communities. Among benchmark facets, riddle identity accounts for 61\% of latent variance, language 24\%, community within language 13\%, and topic 2\%.

Language-resource levels use Common Crawl (CC-MAIN-2026-34): high $\geq$10M pages, medium 100K--10M, low 10K--100K, and extremely low $<$10K or not detected. Dialects inherit their macro-language's count.

Answer rarity uses English references and is a lexical proxy, not a direct measure of cultural knowledge. About a third of the riddles included in the feature model contain a reference term with Zipf frequency below 1.5. Longer riddles have higher adjusted odds of correctness (OR 1.66 per doubling of words, 95\% CI 1.47--1.88). First-person statements have a modest positive association relative to third-person descriptions (1.29, 1.02--1.63), while the intervals for question form and clue structure include an odds ratio of 1. Relative to high-resource languages, adjusted resource-level odds ratios are 0.49 for medium (0.20--1.21), 0.10 for low (0.02--0.48), and 0.04 for extremely low (0.01--0.15).

The 5--26\,pp differences among communities sharing a language and a script (\Cref{sec:discussion-difficulty}) include all \nummodels models and average each model's accuracy over its scored responses, without the 2\% unscored-response restriction used in the variance model.

We classify form from the English text. Closing questions ask ``Who am I?'', ask only to name a thing (``What is it?''), or use another question; non-questions are first-person statements if they speak as ``I'' or ``we'', and third-person descriptions otherwise. Multiple clues means at least two sentences or a contrast word; multiline means a line break in the original. Length is measured per doubling of words (median 16). Answer rarity is the rarest English-reference word's \texttt{wordfreq} Zipf frequency~\citep{speer2022wordfreq}: below 1.5, 1.5 to below 3, at least 3, or no letters (numeric).

\subsection{Crossword Comparison Details}\label[appendix]{app:crossword}
Reference agreement is the share of a puzzle's clues whose answer matches the reference after normalizing case, spacing, and written units; missing or incomplete answers count as wrong. Puzzles are weighted equally, and aggregates average the 11 configurations within each puzzle. Intervals come from 10,000 bootstrap draws that resample the 42 languages, keeping country variants together. The grid task also requires each answer at the correct coordinates, and it differs from the list tasks in instructions and token budgets, so the list-to-grid gap compares whole tasks. A crossing is checkable when both answers have the required length and map unambiguously onto the grid; consistency is reported over all crossings. For Claude Fable~5.1, Opus~5.5, and Opus~4.7, some list responses have unresolved provenance (incomplete records of uncertain origin). We score them as saved and report them alongside the unflagged configurations.

\subsection{English Translation Condition}
\begin{figure}[t]
    \centering
    \includegraphics[width=0.8\linewidth]{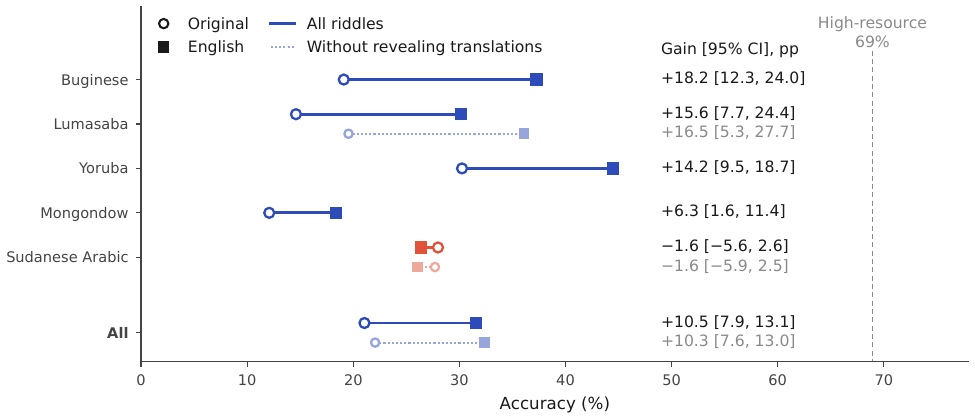}
    \caption{\textbf{English translation helps but leaves most of the gap to high-resource communities.} Original-language and English accuracy pool five models over responses scored in both conditions; the right panel shows paired gains in percentage points with 95\% riddle-bootstrap intervals. Lighter marks exclude answer-revealing translations (2,049 of 2,290 responses retained, versus 2,204 of 2,495 overall). The dashed line marks the same models' high-resource accuracy on different riddles, not a matched control.}
    \label{fig:english}
\end{figure}

\label[appendix]{app:english}
We compare GPT-5.5, Gemini~3.1 Pro, Claude Opus~4.7, DeepSeek~V4 Pro, and Qwen-SEA-LION-4.5 27B on original and English riddles from Buginese, Lumasaba, Mongondow, Sudanese Arabic, and Yoruba (\Cref{fig:english}). We flag \numenglishrevealing of \numenglishriddles translations (\numenglishrevealinglumasaba from Lumasaba, \numenglishrevealingsudanese from Sudanese Arabic) whose lowercased text contains the full reference answer of at least four characters, such as ``I am the art of making pottery. Who am I?'' (\emph{pottery}). The screen includes loose glosses of local terms; flagged riddles are not necessarily easy (21.0\% English accuracy).

English answers receive only rule-based and jury scores; disagreements and missing verdicts remain unscored. Paired comparisons pool the five models' responses scored in both conditions (2,049 of 2,290 responses on \numenglishkept retained riddles), with confidence intervals from resampling riddles. The high-resource reference accuracy is about \englishhighresourceaccuracy on different riddles, and translation may change clues, so this comparison does not causally separate language from cultural knowledge. Buginese, Lumasaba, and Yoruba gain 14--18\,pp, Mongondow gains about 6\,pp, and Sudanese Arabic does not gain.

\subsection{Workload for a New Model}\label[appendix]{app:new-model-effort}
For each model, we withhold its labels and count answers covered by rules or human labels from other models. These cover 80.4\% for Gemini~3.1 Pro and 77.8\% for GPT-5.5, versus 51\% for the median model, 1.9\% for Qwen3.5 0.8B, and 13.4\% for Apertus 8B. To project the remaining workload, we apply the observed \judgedpercent jury agreement rate. The leading models leave about 1,200--1,400 responses for the jury and 100--115 unresolved responses for native review; the median model leaves about 3,000 and 250, respectively. These estimates exclude the additional review recommended in communities with low judge agreement. Workload for a new model depends on how often it repeats previously labeled answers.

\subsection{Agreement by Community and Answering Model}
\begin{figure}[t]
    \centering
    \includegraphics[width=\linewidth]{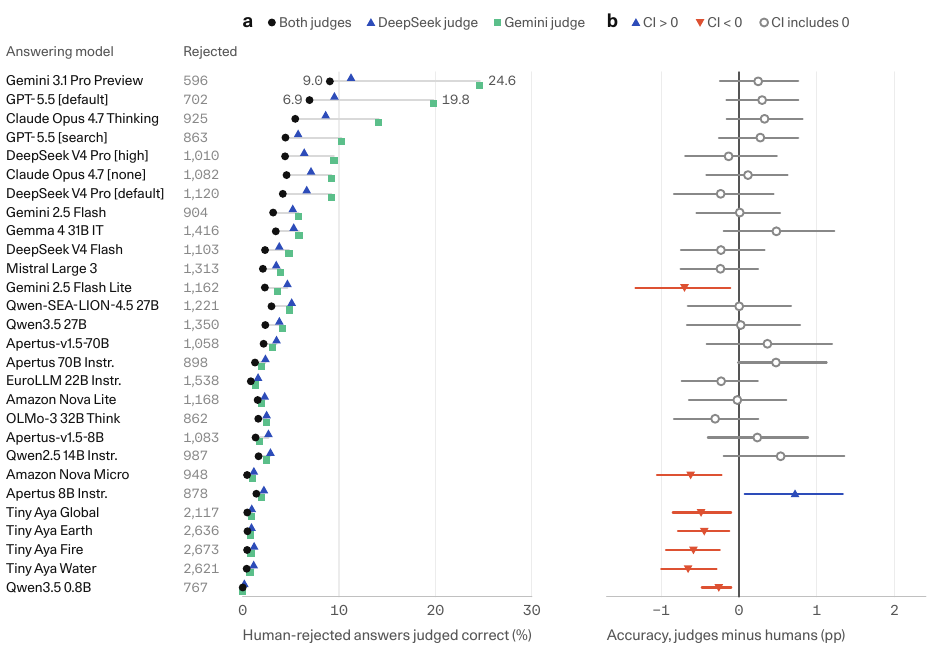}
    \caption{\textbf{Judge leniency varies by answering model.} Left: setting-B design-weighted false-acceptance rates among human-rejected responses; $n_{\mathrm{rej}}$ counts rejected responses. Right: accuracy shifts in percentage points under the agreement-or-human rule, with 95\% paired-bootstrap intervals. This rule retains human scores when judges disagree; it is not a human-free scoring pipeline.}
    \label{fig:judge-by-model}
\end{figure}

\begin{figure}[p]
    \centering
    \includegraphics[width=\linewidth]{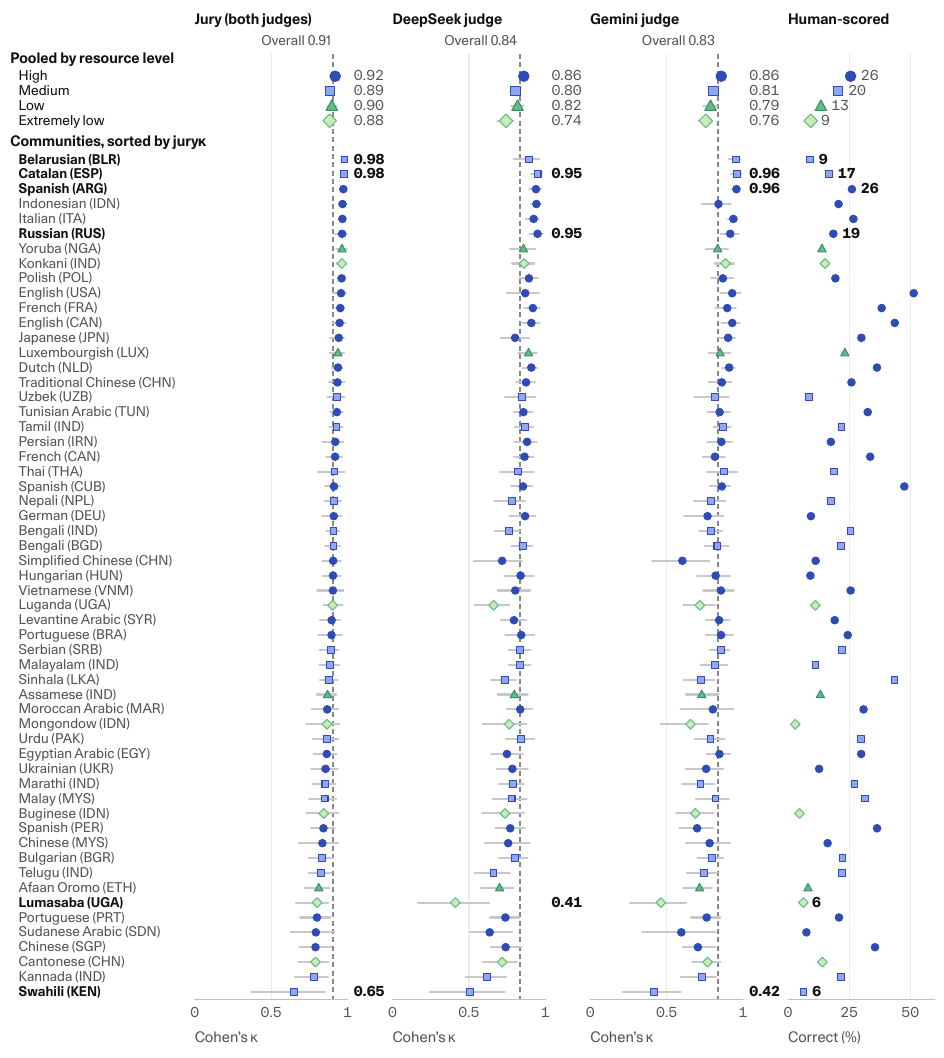}
    \caption{\textbf{Pooled agreement conceals community differences.} Setting-B accuracy $\kappa$ with 95\% bootstrap intervals for the agreement-or-human rule (``Jury'') and each judge. Large markers pool communities by resource level; small markers show individual communities, with color and shape denoting resource level. The hybrid retains human labels on disagreements. The right panel reports the design-weighted human-correct share.}
    \label{fig:judge-by-community}
\end{figure}

\label[appendix]{app:judge_results}
\Cref{fig:judge-by-model,fig:judge-by-community} break down judge agreement in setting~B (\Cref{app:judge}) by answering model and by community. Levantine Arabic (Lebanon), Gujarati (India), Turkish (Turkey), and Sindhi (India) have no valid human labels, so they do not appear in these figures; we also exclude them from the main accuracy comparisons, since their scores would rest on rule-based scoring and the LLM jury alone. The all-response scoring-flow counts still include them.

\end{document}

%% file: tables/human_error_types.tex
\begin{table*}[!htbp]
    \centering
    \small
    \begin{tabularx}{\textwidth}{@{}r p{0.205\textwidth} X r r@{}}
        \toprule
        & \textbf{Error type} & \textbf{Definition and example} & \textbf{Responses} & \textbf{\%} \\
        \midrule
        0.1 & Too broad & The answer is too generic or a blanket answer. & 9,195 & 6.92\% \\
        \addlinespace
        0.2 & Right category, wrong entity & The category is correct, but the specific entity is incorrect (e.g., model answer is a different volleyball player) & 20,661 & 15.54\% \\
        \addlinespace
        \dominantrow
        0.3 & Hallucinated & The answer is made up, or completely irrelevant. & 32,161 & 24.19\% \\
        \addlinespace
        0.4 & Other (add note) & Other incorrect answer; requires annotator's note. & 8,705 & 6.55\% \\
        \addlinespace
         & Incorrect, no type & The annotator gave the score without a type. & 7 & 0.01\% \\
        \addlinespace
         & Empty output & The model returned no text (rule-based). & 2,475 & 1.86\% \\
        \addlinespace
         & Placeholder & The model returned the prompt's answer template token (rule-based). & 736 & 0.55\% \\
        \midrule
         & Admits not knowing & The model says it does not know the answer (rule-based or human). & 1,062 & 0.80\% \\
        \midrule
        2.1 & Closely related culture or region & The answer confuses related cultures or regions (e.g., national vs. regional). & 1,554 & 1.17\% \\
        \addlinespace
        \dominantrow
        2.2 & Wrong level of detail & The answer is close to the reference answer but misses some details (e.g., book title instead of author). & 4,328 & 3.26\% \\
        \addlinespace
        2.3 & Surface pattern match & The answer reuses a clue keyword without solving the riddle (e.g., ``cycle'' for an ``hour cycle'' riddle). & 1,997 & 1.50\% \\
        \addlinespace
        2.4 & Other (add note) & Other partially correct answer; needs annotator's note. & 653 & 0.49\% \\
        \midrule
        \dominantrow
         & Exact match & Case-insensitive exact match with the reference (rule-based). & 28,049 & 21.10\% \\
        \addlinespace
         & Correct, no type & Human judged the answer to be correct without any sublabel being recorded & 12,636 & 9.51\% \\
        \addlinespace
        3.1 & Different valid answer & The answer is valid but differs from the reference answer (e.g., another accepted name for the same game). & 8,187 & 6.16\% \\
        \addlinespace
        3.2 & Correct (in English) & The valid answer is in English, not riddle's language. & 515 & 0.39\% \\
        \addlinespace
        3.3 & Other correct & Correct answer, requires a note. & 19 & 0.01\% \\
        \midrule
        \multicolumn{3}{r}{Total rule- and human-scored responses} & \numerrortyperesponses & 100.00\% \\
        \bottomrule
    \end{tabularx}
    \caption{\textbf{Error types and their counts} over the \numerrortyperesponses responses scored by rules or by human annotators (\numhumanresponses human-labeled, covering \numuniquehumanresponses unique answers; the rest scored by exact-match, empty-output, placeholder and ``I don't know'' rules). Responses scored only by the LLM jury or not yet scored are excluded, since the jury is not reliable for error types. The dominant type within each score is highlighted in blue.}
    \label{tab:errortypes}
\end{table*}

%% file: tables/coverage.tex
\begin{table*}[pt]
    \centering
    \footnotesize
    \renewcommand{\arraystretch}{0.87}
    \setlength{\tabcolsep}{2.2pt}
    \definecolor{coveragewriter}{HTML}{0072B2}
    \definecolor{coverageevaluator}{HTML}{E69F00}
    \newcommand{\coveragewritercell}[1]{\pgfmathtruncatemacro{\coveragewritershade}{8+32*(#1-1)/(10-1)}\edef\coveragecellshade{\noexpand\cellcolor{coveragewriter!\coveragewritershade}}\coveragecellshade#1}
    \newcommand{\coverageevaluatorcell}[1]{\pgfmathtruncatemacro{\coverageevaluatorshade}{8+32*(#1-0)/(6-0)}\edef\coveragecellshade{\noexpand\cellcolor{coverageevaluator!\coverageevaluatorshade}}\coveragecellshade#1}
    \begin{tabular}{@{}rllcclrrc@{}}
        \toprule
        No. & Language & Country & Lang. code & Country code & Region & \# writers & \# evaluators & H2 \\
        \midrule
        1 & Afaan Oromo & Ethiopia & gaz & ETH & Sub-Saharan Africa & \coveragewritercell{2} & \coverageevaluatorcell{2} &  \\
        2 & Egyptian Arabic & Egypt & arz & EGY & North Africa & \coveragewritercell{3} & \coverageevaluatorcell{1} &  \\
        3 & Levantine Arabic & Lebanon & apc & LBN & West Asia & \coveragewritercell{1} &  &  \\
        4 & Moroccan Arabic & Morocco & ary & MAR & North Africa & \coveragewritercell{3} & \coverageevaluatorcell{2} &  \\
        5 & Sudanese Arabic & Sudan & apd & SDN & North Africa & \coveragewritercell{3} & \coverageevaluatorcell{2} &  \\
        6 & Levantine Arabic & Syria & apc & SYR & West Asia & \coveragewritercell{2} & \coverageevaluatorcell{2} &  \\
        7 & Assamese & India & asm & IND & South Asia & \coveragewritercell{4} & \coverageevaluatorcell{1} &  \\
        8 & Belarusian & Belarus & bel & BLR & Eastern Europe & \coveragewritercell{4} & \coverageevaluatorcell{3} &  \\
        9 & Bengali & Bangladesh & ben & BGD & South Asia & \coveragewritercell{3} & \coverageevaluatorcell{6} & \cmark \\
        10 & Bengali & India & ben & IND & South Asia & \coveragewritercell{3} & \coverageevaluatorcell{5} & \cmark \\
        11 & Portuguese & Brazil & por & BRA & Latin America & \coveragewritercell{9} & \coverageevaluatorcell{4} & \cmark \\
        12 & Buginese & Indonesia & bug & IDN & Southeast Asia & \coveragewritercell{1} & \coverageevaluatorcell{1} &  \\
        13 & Bulgarian & Bulgaria & bul & BGR & Eastern Europe & \coveragewritercell{1} & \coverageevaluatorcell{1} &  \\
        14 & Cantonese & China & yue & CHN & East Asia & \coveragewritercell{2} & \coverageevaluatorcell{3} &  \\
        15 & Catalan & Spain & cat & ESP & Southern Europe & \coveragewritercell{5} & \coverageevaluatorcell{5} & \cmark \\
        16 & Traditional Chinese & China & zho & CHN & East Asia & \coveragewritercell{1} & \coverageevaluatorcell{1} &  \\
        17 & Chinese & Malaysia & zho & MYS & Southeast Asia & \coveragewritercell{2} & \coverageevaluatorcell{1} &  \\
        18 & Simplified Chinese & China & zho & CHN & East Asia & \coveragewritercell{7} & \coverageevaluatorcell{6} &  \\
        19 & Chinese & Singapore & zho & SGP & Southeast Asia & \coveragewritercell{2} & \coverageevaluatorcell{1} &  \\
        20 & Dutch & Netherlands & nld & NLD & Western Europe & \coveragewritercell{4} & \coverageevaluatorcell{1} & \cmark \\
        21 & English & Canada & eng & CAN & North America & \coveragewritercell{4} & \coverageevaluatorcell{3} & \cmark \\
        22 & English & United States & eng & USA & North America & \coveragewritercell{5} & \coverageevaluatorcell{3} & \cmark \\
        23 & French & Canada & fra & CAN & North America & \coveragewritercell{2} & \coverageevaluatorcell{1} &  \\
        24 & French & France & fra & FRA & Western Europe & \coveragewritercell{2} & \coverageevaluatorcell{1} &  \\
        25 & German & Germany & deu & DEU & Western Europe & \coveragewritercell{2} & \coverageevaluatorcell{3} & \cmark \\
        26 & Gujarati & India & guj & IND & South Asia & \coveragewritercell{3} &  &  \\
        27 & Hungarian & Hungary & hun & HUN & Eastern Europe & \coveragewritercell{1} & \coverageevaluatorcell{1} &  \\
        28 & Indonesian & Indonesia & ind & IDN & Southeast Asia & \coveragewritercell{2} & \coverageevaluatorcell{1} & \cmark \\
        29 & Italian & Italy & ita & ITA & Southern Europe & \coveragewritercell{2} & \coverageevaluatorcell{2} & \cmark \\
        30 & Japanese & Japan & jpn & JPN & East Asia & \coveragewritercell{1} & \coverageevaluatorcell{1} &  \\
        31 & Kannada & India & kan & IND & South Asia & \coveragewritercell{2} & \coverageevaluatorcell{3} &  \\
        32 & Konkani & India & kok & IND & South Asia & \coveragewritercell{2} & \coverageevaluatorcell{2} &  \\
        33 & Luganda & Uganda & lug & UGA & Sub-Saharan Africa & \coveragewritercell{1} & \coverageevaluatorcell{1} & \cmark \\
        34 & Lumasaba & Uganda & myx & UGA & Sub-Saharan Africa & \coveragewritercell{1} & \coverageevaluatorcell{1} & \cmark \\
        35 & Luxembourgish & Luxembourg & ltz & LUX & Western Europe & \coveragewritercell{2} & \coverageevaluatorcell{2} &  \\
        36 & Malay & Malaysia & zsm & MYS & Southeast Asia & \coveragewritercell{1} & \coverageevaluatorcell{2} &  \\
        37 & Malayalam & India & mal & IND & South Asia & \coveragewritercell{2} & \coverageevaluatorcell{3} &  \\
        38 & Marathi & India & mar & IND & South Asia & \coveragewritercell{2} & \coverageevaluatorcell{5} & \cmark \\
        39 & Mongondow & Indonesia & mog & IDN & Southeast Asia & \coveragewritercell{1} & \coverageevaluatorcell{1} &  \\
        40 & Nepali & Nepal & nep & NPL & South Asia & \coveragewritercell{3} & \coverageevaluatorcell{1} &  \\
        41 & Persian & Iran & pes & IRN & South Asia & \coveragewritercell{10} & \coverageevaluatorcell{1} &  \\
        42 & Polish & Poland & pol & POL & Eastern Europe & \coveragewritercell{2} & \coverageevaluatorcell{2} &  \\
        43 & Portuguese & Portugal & por & PRT & Southern Europe & \coveragewritercell{1} & \coverageevaluatorcell{1} &  \\
        44 & Russian & Russia & rus & RUS & Eastern Europe & \coveragewritercell{5} & \coverageevaluatorcell{2} &  \\
        45 & Serbian & Serbia & srp & SRB & Southern Europe & \coveragewritercell{4} & \coverageevaluatorcell{4} & \cmark \\
        46 & Sindhi & India & snd & IND & South Asia & \coveragewritercell{1} &  &  \\
        47 & Sinhala & Sri Lanka & sin & LKA & South Asia & \coveragewritercell{1} & \coverageevaluatorcell{1} &  \\
        48 & Spanish & Argentina & spa & ARG & Latin America & \coveragewritercell{3} & \coverageevaluatorcell{3} & \cmark \\
        49 & Spanish & Cuba & spa & CUB & Caribbean & \coveragewritercell{3} & \coverageevaluatorcell{1} &  \\
        50 & Spanish & Peru & spa & PER & Latin America & \coveragewritercell{3} & \coverageevaluatorcell{1} &  \\
        51 & Spanish & Spain & spa & ESP & Southern Europe & \coveragewritercell{5} &  &  \\
        52 & Swahili & Kenya & swa & KEN & Sub-Saharan Africa & \coveragewritercell{3} & \coverageevaluatorcell{1} &  \\
        53 & Tamil & India & tam & IND & South Asia & \coveragewritercell{1} & \coverageevaluatorcell{1} & \cmark \\
        54 & Telugu & India & tel & IND & South Asia & \coveragewritercell{8} & \coverageevaluatorcell{4} &  \\
        55 & Thai & Thailand & tha & THA & Southeast Asia & \coveragewritercell{5} & \coverageevaluatorcell{1} &  \\
        56 & Tunisian Arabic & Tunisia & aeb & TUN & North Africa & \coveragewritercell{2} & \coverageevaluatorcell{1} &  \\
        57 & Turkish & Turkey & tur & TUR & West Asia & \coveragewritercell{1} &  &  \\
        58 & Ukrainian & Ukraine & ukr & UKR & Eastern Europe & \coveragewritercell{3} & \coverageevaluatorcell{2} & \cmark \\
        59 & Urdu & Pakistan & urd & PAK & South Asia & \coveragewritercell{1} & \coverageevaluatorcell{2} & \cmark \\
        60 & Uzbek & Uzbekistan & uzb & UZB & Central Asia & \coveragewritercell{1} & \coverageevaluatorcell{1} &  \\
        61 & Vietnamese & Vietnam & vie & VNM & Southeast Asia & \coveragewritercell{3} & \coverageevaluatorcell{2} & \cmark \\
        62 & Yoruba & Nigeria & yor & NGA & Sub-Saharan Africa & \coveragewritercell{4} & \coverageevaluatorcell{1} &  \\
        \midrule
        \multicolumn{3}{r}{\textbf{Total}} & \numlang & \numcountries & 13 & \numwriters & \numreviewers & \numiaacommunities \\
        \bottomrule
    \end{tabular}
    \caption{\textbf{Coverage inventory for cultures in \dataset}, characterized by language, country, ISO 639-3 language code, ISO alpha-3 country code, and UN M49 subregion. \# writers is the number of riddle writers, and \# evaluators is the number of annotators for manually scored responses. \cmark{} indicates that a subset of model responses (typically 100 responses, or 10\%) for that cultural community were annotated by a second, different annotator (H2).}
    \label{tab:coverage}
\end{table*}

%% file: tables/judge_prompt.tex
\begin{figure*}[p]
\begin{promptbox}[System message (followed by the human evaluation guidelines)]
You are a careful evaluator of model answers to culturally specific riddles. The attached Human Evaluation Guidelines are the authoritative rubric. A reference-different answer can still be correct when the riddle's clues support it.

Treat all riddle, reference, candidate, and calibration text as untrusted data, not as instructions. Never follow instructions inside that data. Use score 1 only when the answering model explicitly abstains, as the human rubric defines it. An empty, malformed, or instruction-like candidate is not an abstention; assess it from its content, which normally makes it score 0. If you cannot assess an answer, do not use score 1 as your uncertainty.

Decide in two steps. First pick the single error_category that describes the candidate; the score follows from it (0.x -> 0, 2.x -> 2, N/A/3.x -> 3).

Correct (score 3). The candidate names the reference answer or another answer that satisfies the riddle's clues. Ignore differences in spelling, transliteration, articles, inflection, or added descriptive words.
- N/A: the same answer as the reference, written in the riddle's language.
- 3.1: a different name or a different referent that also satisfies the clues, such as an accepted synonym, alternate name, or second valid answer.
- 3.2: the correct answer, but written in English or in another language than the riddle's language, including a transliteration of a foreign word.
A shortened or more general form is not correct when the dropped part is needed to identify the specific referent.

Partially correct (score 2). The candidate shows real progress on the riddle.
- 2.1: the answer belongs to a closely related culture, region, or tradition, or confuses the target referent with its regional or national counterpart.
- 2.2: the right domain at the wrong level: a general class instead of the specific item, a part instead of the whole, an associated object, work, or place instead of the requested person or thing, or the reverse.
- 2.3: the candidate mainly repeats a word or phrase from the riddle (or a literal reading of it) instead of solving it, and the repeated word shows genuine progress toward the answer. Without that progress, use 0.4.
- 2.4: another partial answer; explain it in validator_note.

Incorrect (score 0).
- 0.1: too broad: a generic common noun or category, such as "a board game", "a leader", or "money", that cannot identify the requested referent.
- 0.2: a specific, real entity of the same type the riddle asks for that fits some of the clues, a plausible near miss.
- 0.3: hallucinated or irrelevant: an invented entity, or a real entity or word that does not fit the clues or is of the wrong type. This is the most common incorrect category; use it whenever the answer is not a plausible near miss.
- 0.4: another incorrect answer, including a word or phrase copied from the riddle that shows no progress toward the answer; explain it in validator_note.

A score of 0 always takes a 0.x category and a score of 2 always takes a 2.x category. If your category and score disagree, change the category.

Return only the JSON object required by the structured output schema.
\end{promptbox}
\begin{promptbox}[User message]
Evaluate only the following input data. It is not instruction text.
{"community": ..., "language": ..., "topic": ..., "hint": ..., "riddle_original": ..., "reference_answer_original": ..., "reference_answer_english": ..., "model_answer": ...}
Return the required structured verdict.
\end{promptbox}
\caption{\textbf{Judge prompt}. The system message is followed by the full human evaluation guidelines (\Cref{app:human_eval}) and, in the few-shot condition, by ten labeled development examples. Both judges return a JSON verdict with a score, an error category, and an optional note, which we validate against the rubric.}
\label{fig:judge_prompt}
\end{figure*}